\documentclass{article} 
\usepackage{iclr2027_conference,times}
\usepackage{amsmath,amsfonts,bm}

\def\eqref#1{equation~\ref{#1}}

\def\1{\bm{1}}

\DeclareMathAlphabet{\mathsfit}{\encodingdefault}{\sfdefault}{m}{sl}
\SetMathAlphabet{\mathsfit}{bold}{\encodingdefault}{\sfdefault}{bx}{n}

\usepackage{hyperref}
\usepackage{url}
\usepackage{booktabs}
\usepackage{tabularx}
\usepackage{array}
\usepackage{booktabs}
\usepackage{graphicx}
\usepackage{wrapfig}
\usepackage[table]{xcolor}
\usepackage{subcaption}
\usepackage{multirow}
\usepackage{float}
\usepackage{tcolorbox}
\tcbuselibrary{breakable}
\tcbuselibrary{skins}
\definecolor{myGreen}{rgb}{0, .6, .0}
\definecolor{myblue}{RGB}{230,245,255}
\definecolor{mydarkblue}{rgb}{0.68, 0.85, 1.0}
\definecolor{mydarkblue2}{rgb}{0,0.08,0.45}
\definecolor{mydarkblue3}{RGB}{151,204,255}
\definecolor{cvprblue}{rgb}{0.21,0.49,0.74}
\definecolor{oxfordblue}{RGB}{0,33,71}
\definecolor{oxfordroyalblue}{RGB}{29,66,166}

\hypersetup{
    colorlinks,
    breaklinks,
    anchorcolor=darkblue,
    citecolor=myGreen
}

\title{Is Better Teacher Supervision Enough? \\
Unlocking Student-side Learning in Multimodal On-Policy Distillation}

\author{Siyuan Liu\textsuperscript{1} \qquad
Kanghui Tian\textsuperscript{2} \qquad
Yue Duan\textsuperscript{1} \qquad
Yutao He\textsuperscript{1} \qquad \\
\textbf{Shangdong Yang}\textsuperscript{\textbf{3}} \qquad
\textbf{Jian Zhang}\textsuperscript{\textbf{1}}\thanks{Corresponding author} \qquad
\textbf{Yinghuan Shi}\textsuperscript{\textbf{1}}\footnotemark[2] \\
\textsuperscript{1}Nanjing University \,
\textsuperscript{2}Fudan University \,
\textsuperscript{3}Nanjing University of Posts and Telecommunications \\
\texttt{syliu@smail.nju.edu.cn, \{zhang.jian,syh\}@nju.edu.cn}
}

\iclrfinalcopy 
\begin{document}
\maketitle
\fancyhead{}
\begin{figure}[h]
    \centering
    \begin{minipage}{0.93\linewidth}
        \centering
        \refstepcounter{subfigure}
        \label{fig:overview}
        \includegraphics[width=\linewidth]{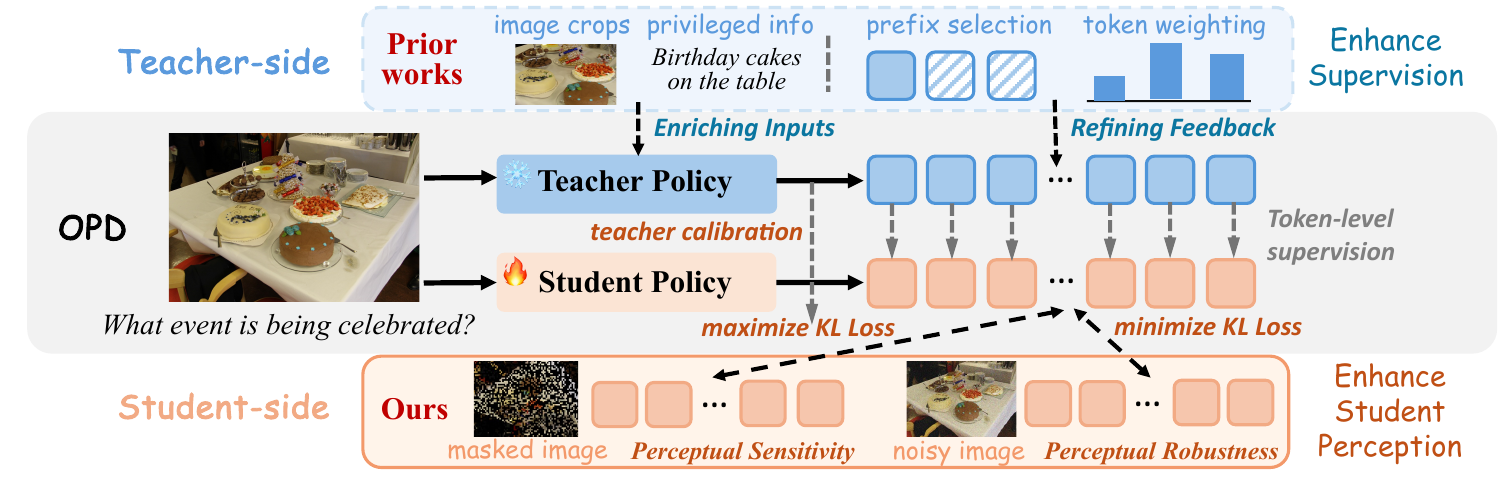}
    \end{minipage}
    \par\vspace{10pt}
    \begin{minipage}{0.93\linewidth}
        \centering
        \refstepcounter{subfigure}
        \label{fig:details}
        \includegraphics[width=\linewidth]{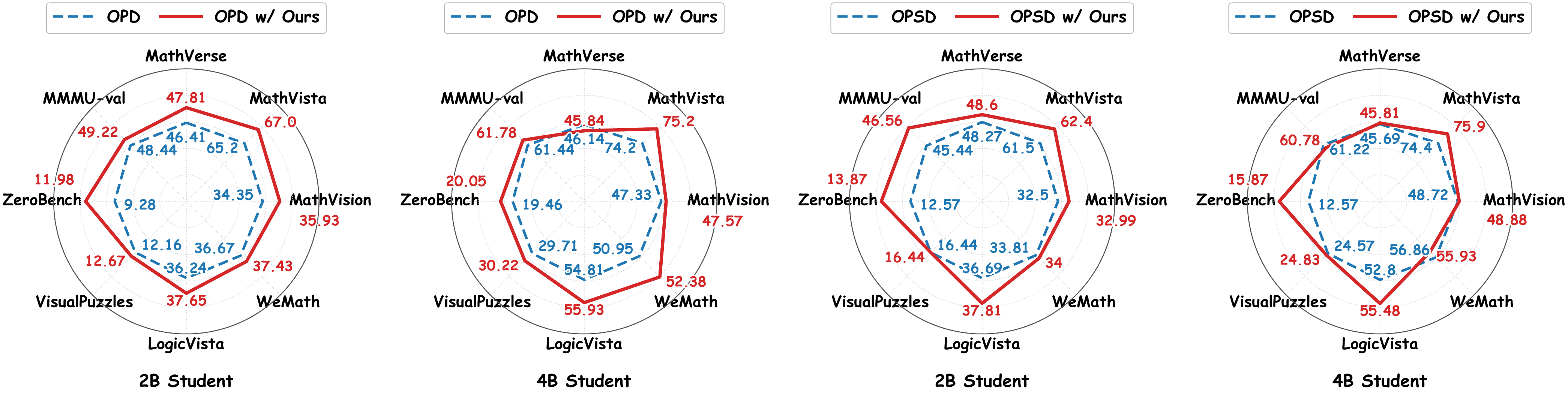}
    \end{minipage}
    \caption{\textbf{Beyond enhancing teacher-side supervision:} Previous methods improve the teacher-side signal, while S-OPD adds two student-side objectives to enhance perception. This improves performance across model scales (2B and 4B) and distillation paradigms (OPD and OPSD).}
    \label{fig:framework}
\end{figure}
\begin{abstract}
On-policy distillation (OPD) improves reasoning by providing token-level supervision from a teacher on a student's own trajectories. Existing methods primarily focus on enhancing this teacher-side guidance (e.g., by enriching teacher inputs and refining teacher feedback), yet we find that limited student perception is another critical bottleneck in multimodal OPD. By providing oracle visual facts, the performance of OPD-trained students can still be substantially improved for both weak and strong teachers. To address this bottleneck, we propose S-OPD, a simple multimodal on-policy distillation framework that explicitly strengthens student perceptual learning through two objectives. Specifically, \textit{Teacher-calibrated Policy Contrast} separates student policies under original and masked images with teacher-based token-level gating, strengthening the student's reliance on visual evidence during reasoning. \textit{Policy Agreement} aligns student policies under original and noise-perturbed images, further improving perceptual robustness to visual noise. Notably, our method can be seamlessly plugged into existing OPD frameworks, requiring no additional data annotations, model parameters or inference operations. Extensive experiments on eight benchmarks across student scales and distillation paradigms demonstrate consistent performance improvements, with gains of up to 4.25 points on LogicVista.  When combined with existing teacher-side supervision methods, our method can yield further gains. Code is available at \url{https://github.com/Sirilaw/S-OPD}.
\end{abstract}

\section{Introduction}
\label{sec:intro}

Multimodal large language models (MLLMs)~\citep{bai2025qwen3vl,zhu2025internvl3} have demonstrated strong capabilities in solving tasks that require models to integrate visual information with multi-step reasoning, like visual question answering~\citep{liu2024improved,ma2026tools}, geometry problem solving~\citep{lu2021geo3k,lu2024mathvista} and chart-oriented reasoning~\citep{masry2022chartqa, chen2025chartr1}. This motivates efforts to equip compact models for these capabilities with lower deployment costs. On-policy distillation (OPD) ~\citep{gu2024minillm,agarwal2024gkd,lu2025onpolicydistillation} is appealing as it trains a student on its own trajectories using feedback from a strong teacher, helping the student learn better behavior at its own rollout distribution during training. Compared to outcome-based reinforcement learning~\citep{shao2024deepseekmath,guo2025deepseek}, OPD provides dense token-level supervision by scoring student-generated prefixes with the teacher model, allowing intermediate decisions to receive direct supervision. 

\begin{wrapfigure}[25]{r}{0.5\columnwidth}
    \centering
    \vspace{-14pt}
    \begin{subfigure}{\linewidth}
        \centering
        \includegraphics[width=\linewidth]{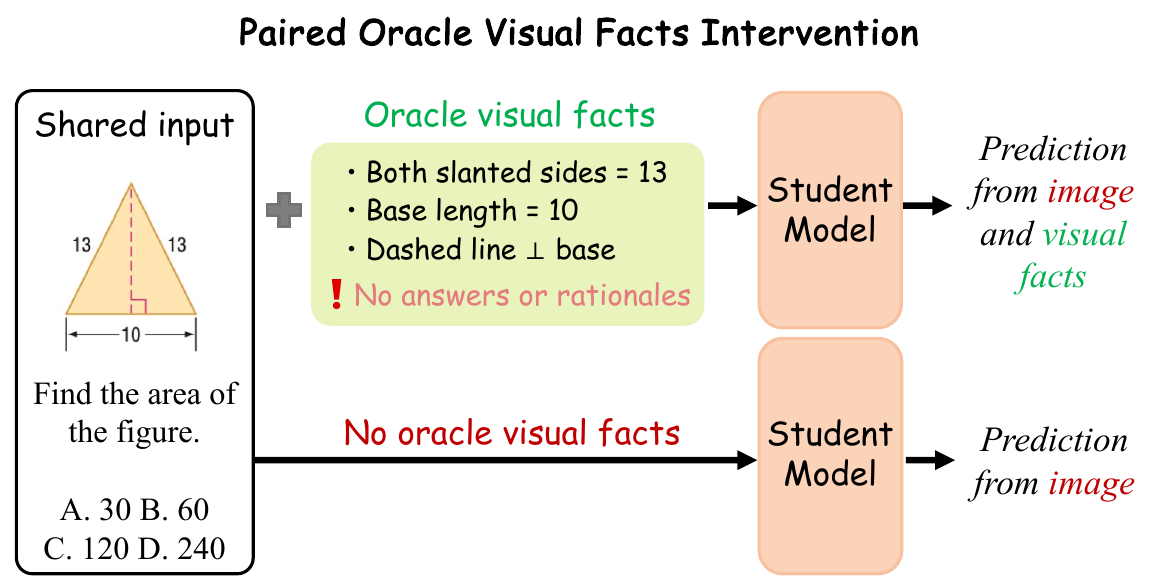}
        \caption{Evaluation pipeline.}
        \label{fig:visual_facts_pip}
    \end{subfigure}
    \par\vspace{2pt}
    \begin{subfigure}{\linewidth}
        \centering
        \includegraphics[width=\linewidth]{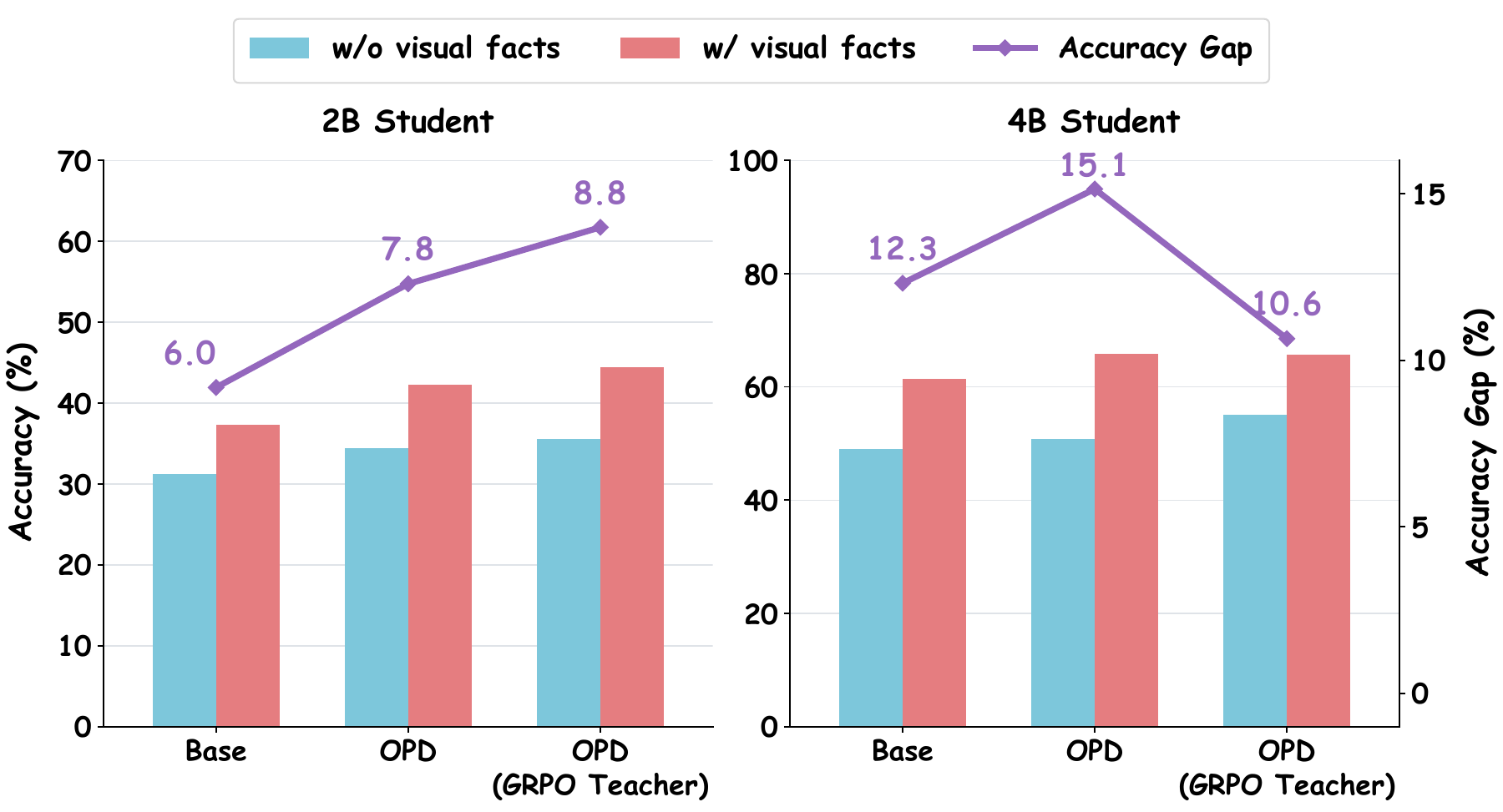}
        \caption{Accuracy and accuracy gaps for 2B and 4B students.}
        \label{fig:visual_facts_acc}
    \end{subfigure}
    \caption{\textbf{Oracle visual facts evaluation on Geometry3K.}}
    \label{fig:visual_facts_pip_acc}
\end{wrapfigure}
Recent efforts to improve OPD primarily focus on two aspects: enriching teacher inputs and refining teacher feedback. The first provides teachers with additional privileged information, such as reference rationales~\citep{zhao2026opsd} and in multimodal settings, image crops~\citep{yuan2026vision} or recoverable visual cues~\citep{tian2026vicur}. The second utilizes differences of teacher predictions conditioned on various visual inputs to reweight the token-level supervision~\citep{liu2026vaopd, liang2026visual} or reconstruct teacher targets~\citep{zhang2026vad}.
Despite these advances, genuine multimodal reasoning requires models to integrate visual perception, a critical component that provides visual evidence, with multi-step textual reasoning~\citep{zhang2024mathverse,tong2024eyes,wang2026papo}. Prior approaches seek to improve the teacher-side guidance that the student receives. Yet this guidance is conveyed through text-token supervision. Students with limited visual perception may struggle to ground the guidance in the visual inputs, limiting the benefits of distillation. This motivates us to ask: \textit{Is student-side visual perception capability another critical bottleneck for multimodal OPD?}

To investigate this question, we evaluate base and OPD-trained students, including Qwen3-VL-2B-Instruct and Qwen3-VL-4B-Instruct~\citep{bai2025qwen3vl}, on the same Geometry3K problems with and without oracle visual facts (Figure~\ref{fig:visual_facts_pip}). These facts describe detailed diagram content without providing answers or reasoning steps, reducing the demand on visual perception (see Section~\ref{sec:method_problem_formulation} for more details). The resulting accuracy gap indicates how much students benefit when the required visual information is explicitly available in textual form. A larger gap suggests that students still struggle to extract task-relevant visual evidence on their own despite training. As shown in Figure~\ref{fig:visual_facts_acc}, we find that substantial gaps persist after OPD, even with stronger teachers~\citep{li2026rethinking} that are trained with GRPO~\citep{shao2024deepseekmath}, suggesting that teacher supervision alone does not sufficiently address students' perceptual limitations.
\looseness=-1

\begin{wrapfigure}[18]{r}{0.5\columnwidth}
    \centering
    \vspace{-10pt}
    \includegraphics[width=\linewidth]{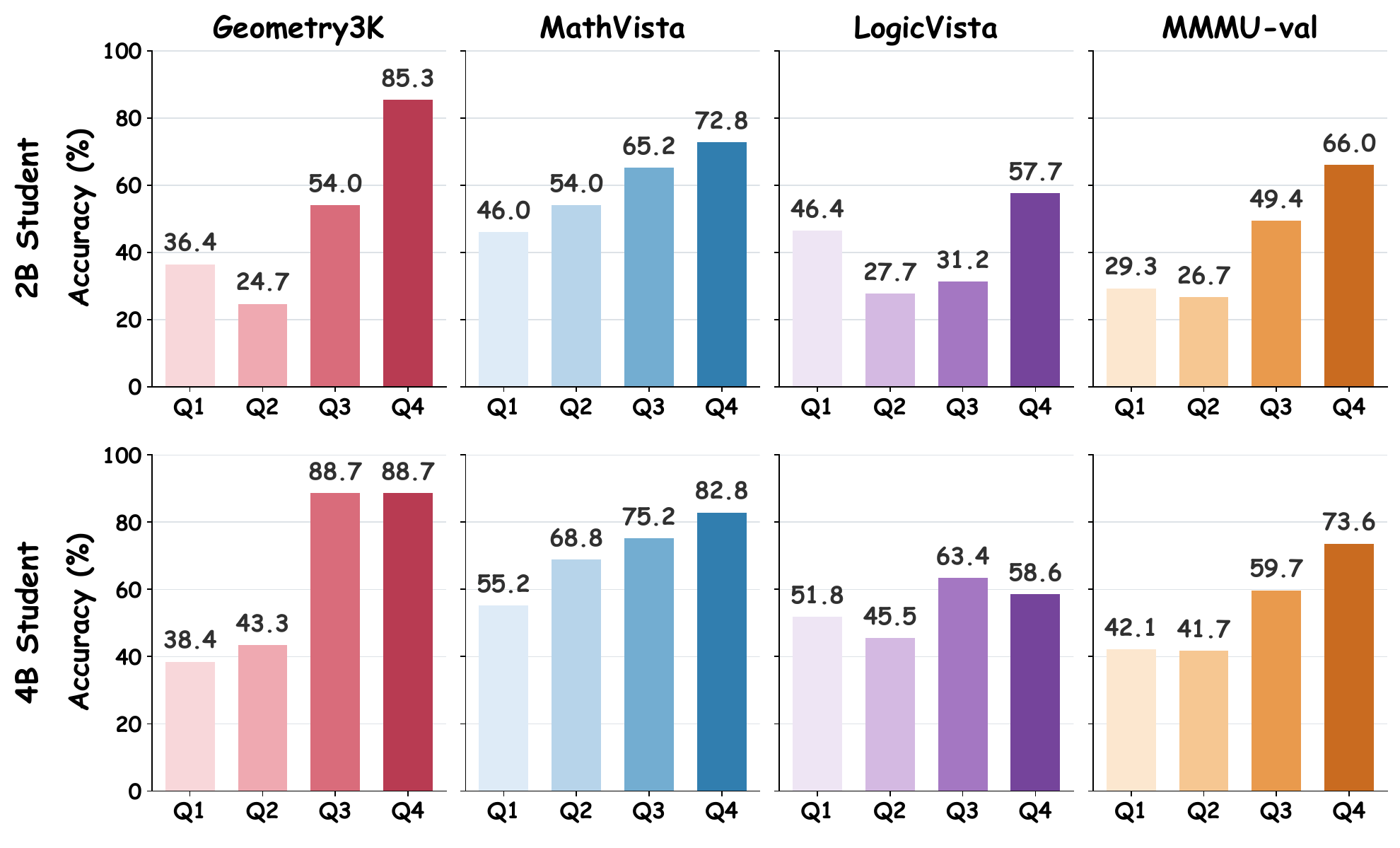}
    \caption{\textbf{Performance across quartiles of estimated KL divergence} between student policies under original and masked images (Q1--Q4, lowest to highest). A clear trend is observed: Higher-KL groups generally achieve higher accuracy.}
    \label{fig:kl_acc_relation}
\end{wrapfigure}

The persistent gap raises a further question: \textit{How does student visual perception influence reasoning performance after OPD on different benchmarks?} To examine this, we systematically analyze four benchmarks covering mathematical, logical and general domains (see Section~\ref{sec:method_problem_formulation} for more details). Specifically, we estimate the KL divergence between two student policies under original and masked images to measure the visual sensitivity of the student. We then divide samples into four quartiles according to this measure and compare model performance across groups. As shown in Figure~\ref{fig:kl_acc_relation}, samples with higher KL divergence--i.e. those exhibiting greater sensitivity to visual input--generally achieve better performance across both student scales. This observation demonstrates a strong positive correlation between the student visual perception and reasoning performance, highlighting the necessity of explicitly strengthening student perceptual learning within OPD.

Building upon our earlier observations, \textit{how can we exploit OPD's token-level learning structure to explicitly strengthen student perception?} We propose \textbf{S-OPD}, a simple multimodal OPD framework that explicitly strengthens student perceptual learning with two KL objectives. Our key idea is to shape student perceptual sensitivity according to whether an input change alters visual evidence: \textit{visual evidence removal should induce a corresponding policy change}, while \textit{perturbations preserving task-relevant content should maintain consistent predictions.} To enhance perceptual sensitivity to evidence removal, Teacher-calibrated Policy Contrast (TPC) maximizes the KL divergence between student policies under original and masked images. To focus this objective on \textit{visually dependent} tokens~\citep{huang2026spotlight}, we use differences in the teacher's predictions under the same two images to gate which tokens KL maximization is applied to. Ablations of different gating strategies in Section~\ref{sec:ablation} demonstrate the importance of this teacher-based selection. However, encouraging policy divergence under evidence removal alone does not constrain the student's response to irrelevant visual variations. We therefore introduce Policy Agreement (PA) as a complementary regularizer that minimizes the KL divergence between student policies under original and noise-perturbed images.

As shown in Figure~\ref{fig:framework}, despite its simplicity, S-OPD achieves consistent performance gains on various reasoning tasks across student scales and distillation paradigms. In particular, methods that enhance teacher-side supervision achieve further performance gains when equipped with S-OPD, highlighting the additional benefits of our student perceptual learning. 

Our contributions are:
\begin{itemize}
    \item We investigate an overlooked student-side visual perception bottleneck in multimodal on-policy distillation, and reveal a positive association between students' perceptual sensitivity and reasoning performance across diverse tasks and models.
    \item We propose S-OPD, a simple multimodal OPD framework that enhances student perceptual learning with two KL objectives targeting perceptual sensitivity and stability. 
    \item Extensive experiments demonstrate consistent performance improvements across multiple student scales, distillation paradigms and benchmarks, with gains reaching 4.25 points on LogicVista. Methods that enhance teacher-side supervision achieve further gains when augmented with S-OPD.
\end{itemize}

\section{Related Work}
\label{sec:related}

\paragraph{On-Policy Distillation.}
Knowledge distillation~\citep{hinton2015distilling} trains a student to match a larger teacher's predictive distribution, while sequence-level distillation~\citep{kim2016sequence} uses teacher-generated sequences as training targets. However, training on fixed or teacher-generated sequences creates a mismatch with inference~\citep{agarwal2024gkd}, where students condition on their own generated prefixes.
On-policy distillation~\citep{gu2024minillm,lu2025onpolicydistillation} addresses this mismatch by having the teacher provide dense token-level supervision on trajectories sampled by the student. Recent work improves this supervision along two dimensions: enriching teacher inputs~\citep{zhao2026opsd,tian2026selfteacherseeprivilegedcontext} to improve teacher advantage and refining teacher feedback~\citep{jin2026entropyopd,jang2026veto,xu2026tip} to improve the reliability of token-level guidance. 
These efforts primarily enhance teacher-side supervision, while S-OPD focuses on student-side learning to help students better exploit the teacher guidance they receive.

\paragraph{Multimodal On-Policy Distillation.}
Recent work improves multimodal supervision from three perspectives: enriching its sources through text-to-vision reasoning transfer~\citep{bousselham2026vold} or heterogeneous teacher capabilities~\citep{yin2026h}; constructing informative teacher--student asymmetry through privileged teacher crops~\citep{yuan2026vision}, recoverable visual cues~\citep{tian2026vicur}, or augmented student views~\citep{li2026s2vopd}; and refining distillation signals through visual interventions~\citep{sun2026vzero} and modality-aware constraints~\citep{bi2026opdv}. Within signal refinement, VA-OPD~\citep{liu2026vaopd} uses teacher visual advantage to allocate supervision across rollouts and token groups
and VAD~\citep{zhang2026vad} reconstructs visually attributable distillation targets. Beyond these advances, S-OPD directly shapes student visual perceptual sensitivity through two cross-view KL objectives, explicitly promoting sensitivity to evidence removal and stability under content-preserving perturbations.

\paragraph{Perceptual Learning for Multimodal Reasoning.}
Prior work has explored multiple training and inference strategies for improving visual perception. Visual contrast uses prediction differences to strengthen visual grounding in decoding~\citep{leng2024mitigatingvcd} and preference optimization~\citep{xie2024vdpo}, while consistency learning promotes the robustness of neural networks in stability training~\citep{zheng2016improving} and semi-supervised learning~\citep{tarvainen2017meanteacher}. In multimodal reinforcement learning, Video-R1~\citep{feng2026video} contrasts temporally ordered and shuffled video inputs to encourage temporal reasoning, and PAPO~\citep{wang2026papo} encourages separation between original- and masked-image policies to strengthen visual perception. Token-selective methods ~\citep{huang2026spotlight} focus optimization on tokens with higher visual dependency.

Unlike prior applications of visual contrast and consistency, S-OPD addresses the student perceptual bottleneck that limits the effectiveness of teacher supervision in OPD. Notably, S-OPD uses the changes in teacher-assigned token probabilities to select tokens for policy contrast. Teacher feedback thus provides both distillation targets and token-level calibration for student perceptual learning.

\section{Method}
\label{sec:method}
\begin{figure}[t]
    \centering
    \begin{minipage}[t]{0.5\linewidth}
        \centering
        \includegraphics[width=\linewidth]{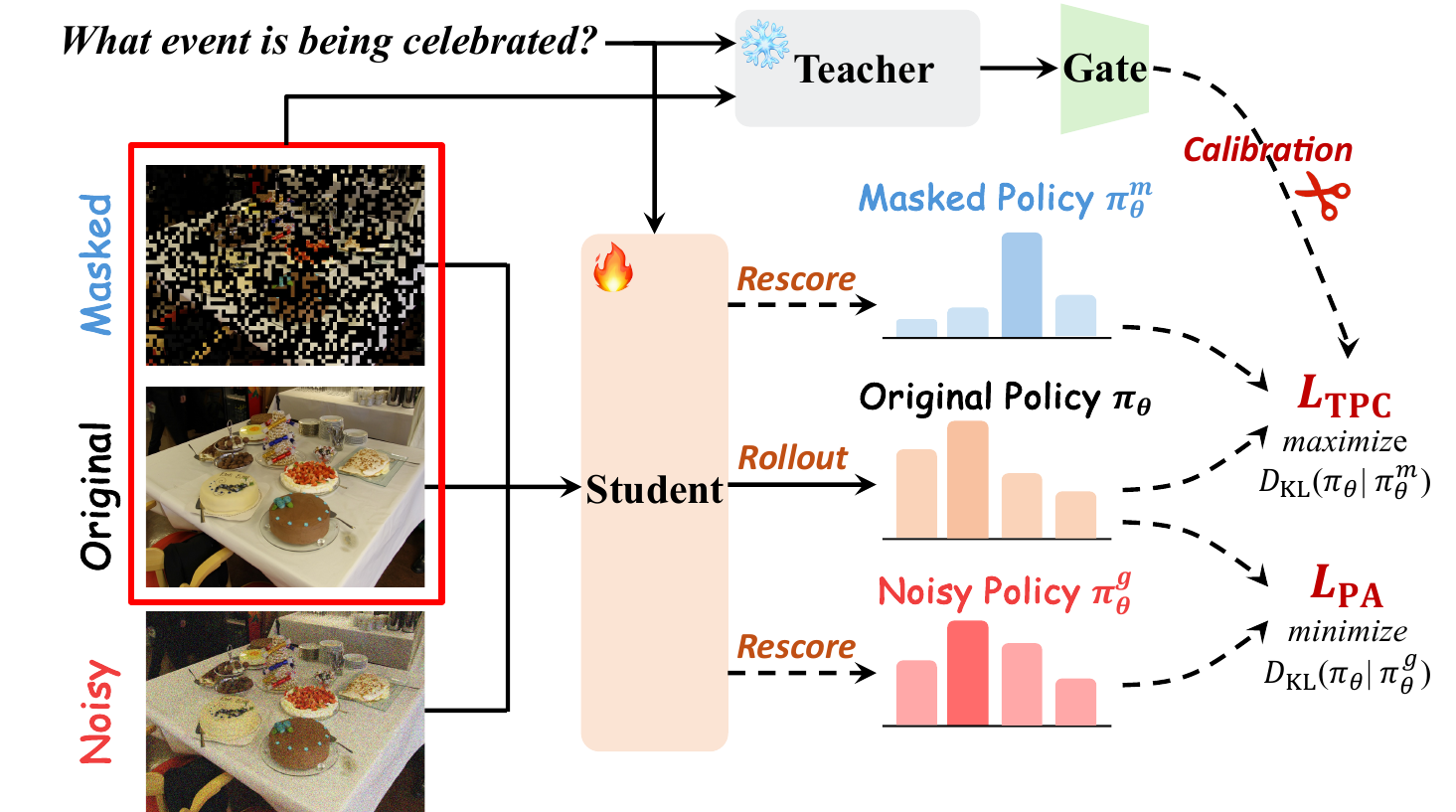}
        \subcaption{Details of two learning objectives in S-OPD.}
        \label{fig:details_of_kl_objective}
    \end{minipage}\hfill
    \begin{minipage}[t]{0.5\linewidth}
        \centering
        \includegraphics[width=\linewidth]{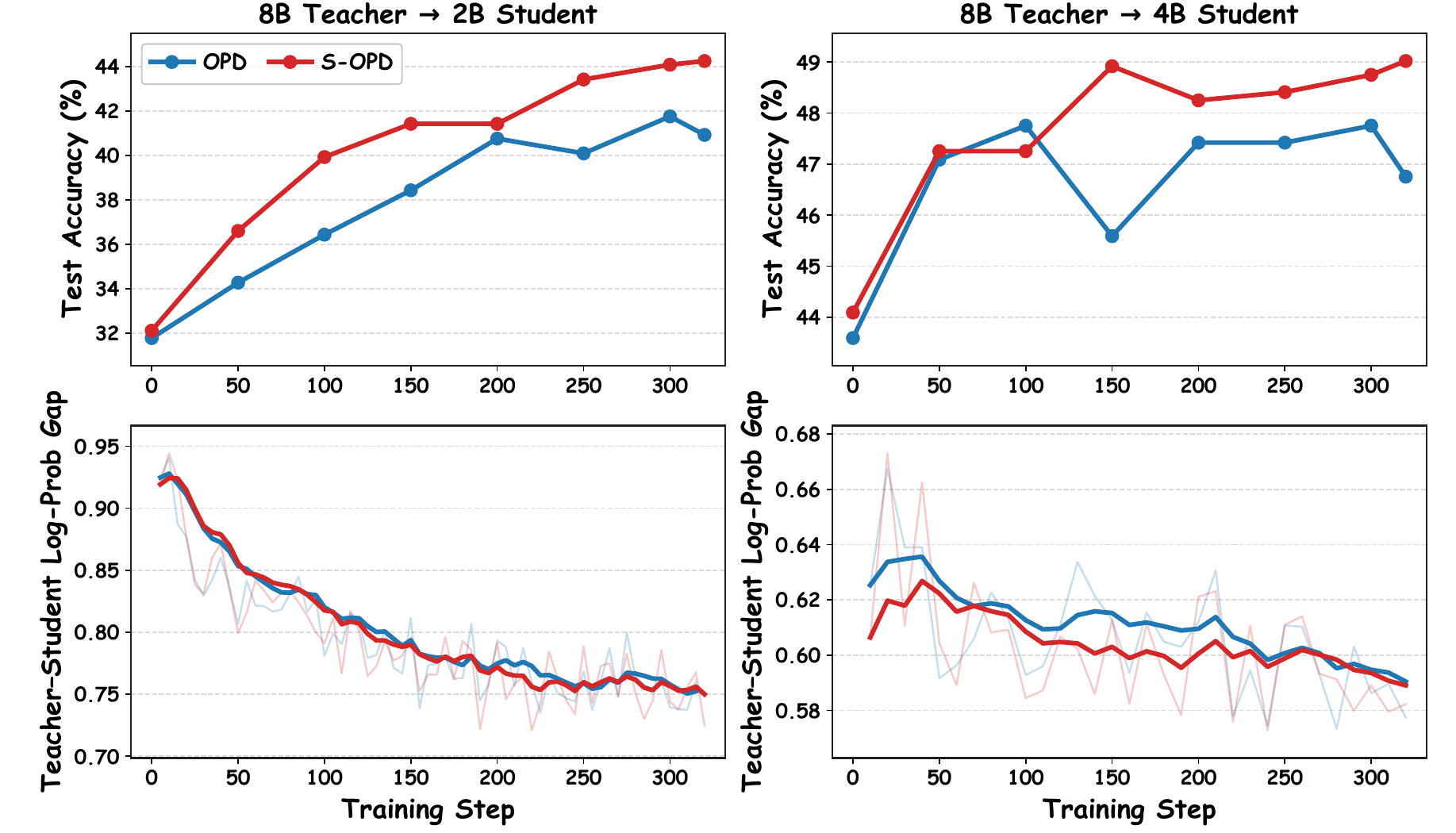}
        \subcaption{Training dynamics of 2B and 4B student models.}
        \label{fig:details-b}
    \end{minipage}
    \caption{(a) Details of S-OPD with original, masked, and noisy images. The teacher compares original and masked images to gate student policy contrast, while policy agreement aligns the student's original and noisy policies. (b) S-OPD achieves higher performance and reduces the teacher-student log-probability gap during training on Geometry3K.}
    \label{fig:method-details}
\end{figure}

We propose \textbf{S-OPD}, a simple framework that explicitly shapes student visual perception sensitivity through two KL objectives. Teacher-Calibrated Policy Contrast (TPC) maximizes the KL divergence between student policies under original and masked images at tokens selected using the teacher's cross-image responses, \textit{strengthening perceptual sensitivity to evidence removal}. Policy Agreement (PA) minimizes the KL divergence between student policies under original and mildly perturbed images, \textit{improving perceptual stability under content-preserving perturbations.} Figure~\ref{fig:framework} illustrates the overall framework and Figure~\ref{fig:method-details} shows the details and training dynamics of our method.

\subsection{Problem Formulation}
\label{sec:method_problem_formulation}

Let $\mathcal{D}$ denote the training distribution of multimodal inputs $x=(q,I)$, where $q$ is a question and $I$ is an image. Let $\pi_\theta$ denote the student policy and $\pi_T$ the frozen teacher policy. For each input, the student samples a response $y=(y_1,\ldots,y_T)\sim\pi_\theta(\cdot\mid x)$, where $T$ is the response length. We define $c_t=(q,I,y_{<t})$ as the context at step $t$, under which both policies are evaluated using the same student-generated prefix.

Following prior OPD studies~\citep{gu2024minillm,lu2025onpolicydistillation}, we train the student to minimize the reverse KL divergence to the teacher:
\begin{equation}
\mathcal{L}_{\mathrm{OPD}}(\theta)
=
\mathbb{E}_{x\sim\mathcal{D},\,y\sim\pi_\theta(\cdot\mid x)}
\left[
\frac{1}{T}
\sum_{t=1}^{T}
D_{\mathrm{KL}}
\left(
\pi_\theta(\cdot\mid c_t)
\,\middle\|\,
\pi_T(\cdot\mid c_t)
\right)
\right].
\label{eq:opd_objective}
\end{equation}
In practice, we estimate the KL divergence using the sampled-token
$k_1$ estimator,
$\log\pi_\theta(y_t\mid c_t)-\log\pi_T(y_t\mid c_t)$,
evaluated on student-sampled tokens.

\paragraph{Bottleneck analysis of multimodal on-policy distillation.}
We first investigate whether student visual perception remains a bottleneck for multimodal OPD. We train Qwen3-VL-2B-Instruct and Qwen3-VL-4B-Instruct~\citep{bai2025qwen3vl} on ViRL39K for 1 epoch with standard OPD~\citep{lu2025onpolicydistillation}. We use Qwen3-VL-8B-Instruct~\citep{bai2025qwen3vl} and its variant trained with GRPO on ViRL39K~\citep{vl-rethinker} as teachers, following the use of post-trained teachers to strengthen supervision~\citep{li2026rethinking}. Base and OPD-trained students are evaluated on the Geometry3K test split with and without oracle visual facts. These facts are constructed by verbalizing official \texttt{diagram\_logic\_form} annotations with fixed templates, without accessing answers or rationales. We insert the facts after the answer choices as a \texttt{Diagram observations} block, keeping all other inputs and decoding settings fixed. Construction and audit details can be found in Appendix~\ref{app:probe_details}. As shown in Figure~\ref{fig:visual_facts_acc}, oracle facts substantially improve accuracy even after OPD with the stronger GRPO teacher, indicating persistent student limitations in visual perception, i.e. the ability to recover visual evidence.

\paragraph{Visual perception sensitivity and reasoning performance.}
We next examine how student visual perception sensitivity relates to reasoning performance on Geometry3K~\citep{lu2021geo3k}, MathVista~\citep{lu2024mathvista}, LogicVista~\citep{xiao2024logicvista}, and MMMU~\citep{yue2024mmmu}, covering mathematical, logical and general reasoning. For each student response generated from the original image, we measure the average token-level KL divergence between student predictive distributions under original and randomly masked images, keeping the question and response prefixes fixed:
\begin{equation}
s(x,y)
=
\frac{1}{T}\sum_{t=1}^{T}
D_{\mathrm{KL}}\!\left(
\pi_\theta(\cdot\mid c_t)
\,\Vert\,
\pi_\theta(\cdot\mid c_t^{m})
\right),
\label{eq:visual_sensitivity}
\end{equation}
where $c_t^{m}$ replaces the original image in $c_t$ with its masked version and higher $s$ indicates greater sensitivity to evidence removal. Within each benchmark and student scale, we divide examples into quartiles by their estimated scores. Figure~\ref{fig:kl_acc_relation} shows that higher-scoring groups generally achieve higher accuracy across both student scales. 

Together, these findings motivate explicitly shaping how students respond to visual evidence during OPD. S-OPD augments teacher--student supervision with student-side objectives that encourage sensitivity to evidence removal and consistency under mild perturbations. To define these objectives, we introduce masked and noisy image views: \begin{equation} 
I^m=\mathcal{M}(I;\rho), \qquad I^g=\operatorname{clip}(I+\epsilon), \quad \epsilon\sim\mathcal{N}(0,\sigma^2\mathbf{I}), 
\label{eq:views} 
\end{equation} 
where $\mathcal{M}$ randomly masks a proportion $\rho$ of image patches, $\sigma$ is the Gaussian noise standard deviation, $\mathbf{I}$ is the identity matrix, and $\operatorname{clip}$ restricts pixels to their valid range. Masking reduces available visual evidence, while mild noise is intended to preserve task content. The corresponding contexts, $c_t^m=(q,I^m,y_{<t})$ and $c_t^g=(q,I^g,y_{<t})$, share the original response, enabling comparisons at the same position. We provide the visualization of masked and noisy images in Appendix~\ref{app:train_details}.

\subsection{Teacher-Calibrated Policy Contrast}
\label{sec:tpc}
To enhance student visual perception, TPC encourages the student to use task-relevant visual evidence. Specifically, it increases the discrepancy between the student's policies under original and masked images at tokens where the teacher indicates that the removed evidence is relevant.

To focus this contrast on tokens affected by masking, we select positions where the teacher assigns greater probability under the original image:
\begin{equation}
\Delta_t^T
=
\log\frac{\pi_T(y_t\mid c_t)}
{\pi_T(y_t\mid c_t^m)},
\qquad
g_t
=
\mathbb{I}\left[\Delta_t^T>\tau\right],
\label{eq:teacher_gate}
\end{equation}
where $\tau$ specifies the required log-probability drop and is set to $0$ by default. Despite using a zero margin, the teacher gate remains selective in practice, activating on only 46.9\% of response tokens; detailed per-rollout statistics are provided in Appendix~\ref{app:sparsity}. This gate uses the teacher's response to masking as a token-level signal of visual dependence~\citep{huang2026spotlight}. At the selected positions, TPC maximizes the KL divergence between the student's original and masked policies, weighted by the teacher's log-probability drop:
\begin{equation}
\mathcal{L}_{\mathrm{TPC}}(\theta)
=
-
\mathbb{E}_{x\sim\mathcal{D},\,
y\sim\pi_\theta(\cdot\mid q,I)}
\left[
\frac{1}{Z}
\sum_{t=1}^{T}
g_t\,\Delta_t^T\,
D_{\mathrm{KL}}
\left(
\pi_\theta(\cdot\mid c_t)
\,\middle\|\,
\pi_\theta(\cdot\mid c_t^m)
\right)
\right],
\label{eq:tpc_objective}
\end{equation}
where $Z=\max(1,\sum_{t=1}^{T}g_t)$ normalizes by the number of selected tokens. In practice, we use the sampled-token $k_3$ estimator, $r_t^m-\log r_t^m-1$, where $r_t^m=\pi_\theta(y_t\mid c_t^m)/\pi_\theta(y_t\mid c_t)$, and stop gradients through the masked-image policy. Thus, the teacher determines where contrast is applied, while the student's cross-image discrepancy provides the contrastive learning signal.

\subsection{Policy Agreement}
\label{sec:pa}

To make the student's use of visual evidence robust to local image variations, PA encourages consistent predictions under mild Gaussian perturbations. These perturbations are intended to preserve task-relevant content, so the student's predictions should remain stable despite changes in individual pixel values. Accordingly, PA minimizes the KL divergence between the student's original and noisy policies at the same response prefixes:
\begin{equation}
\mathcal{L}_{\mathrm{PA}}(\theta)
=
\mathbb{E}_{x\sim\mathcal{D},\,
y\sim\pi_\theta(\cdot\mid q,I)}
\left[
\frac{1}{T}
\sum_{t=1}^{T}
D_{\mathrm{KL}}
\left(
\pi_\theta(\cdot\mid c_t)
\,\middle\|\,
\pi_\theta(\cdot\mid c_t^g)
\right)
\right].
\label{eq:pa_objective}
\end{equation}
We implement the KL divergence using the sampled-token $k_3$ estimator, $r_t^g-\log r_t^g-1$, where
$r_t^g=\pi_\theta(y_t\mid c_t^g)/\pi_\theta(y_t\mid c_t)$, and stop gradients through the noisy-image policy. The noisy-image policy thus provides a fixed target for each update, encouraging the student to maintain consistent predictions under perturbations that preserve task-relevant visual evidence.

\subsection{Overall Objective}
\label{sec:objective}

S-OPD combines teacher--student distillation with explicit objectives for visual perception:
\begin{equation}
\mathcal{L}_\mathrm{{S\text{-}OPD}}(\theta)
=
\mathcal{L}_{\mathrm{OPD}}(\theta)
+
\lambda_{\mathrm{TPC}}\mathcal{L}_{\mathrm{TPC}}(\theta)
+
\lambda_{\mathrm{PA}}\mathcal{L}_{\mathrm{PA}}(\theta),
\label{eq:sopd}
\end{equation}
where $\lambda_{\mathrm{TPC}}$ and $\lambda_{\mathrm{PA}}$ control the strengths of policy contrast and agreement. Alongside token-level teacher guidance, these objectives train the student to respond to relevant evidence removal and remain stable under mild visual perturbations.

For each input, S-OPD samples only a single rollout under the original image. Notably, S-OPD can be seamlessly plugged into existing OPD frameworks, requiring no additional model parameters or inference operations. Its additional computation is confined to training, with a modest and practical overhead (see Appendix~\ref{app:computational_cost}).
\looseness=-1

\section{Experiments}
\label{sec:experiments}

Our experiments test whether strengthening student visual perception enables more effective use of teacher supervision in multimodal OPD. We address four questions. \textbf{Q1:} Does S-OPD improve multimodal OPD across student scales and teacher strengths? \textbf{Q2:} Are these gains sustained across various tasks beyond the training domain? \textbf{Q3:} How do TPC and PA contribute to these improvements, and does teacher calibration enhance the effectiveness of student policy contrast? \textbf{Q4:} What additional computational cost does S-OPD introduce?
\looseness=-1

\paragraph{Data.}
For the main experiments, we train on ViRL39K~\citep{vl-rethinker}. We evaluate the models across eight benchmarks covering three capabilities: mathematical reasoning on MathVerse~\citep{zhang2024mathverse}, MathVista~\citep{lu2024mathvista}, MathVision~\citep{wang2024mathvision}, and WeMath~\citep{qiao2024wemath}; logical reasoning on LogicVista~\citep{xiao2024logicvista} and VisualPuzzles~\citep{song2025visualpuzzles}; and general visual reasoning on ZeroBench-sub~\citep{roberts2025zerobench} and MMMU-val~\citep{yue2024mmmu}. Geometry3K~\citep{lu2021geo3k} is reserved for ablation studies due to its compact size and focused geometry domain.
\looseness=-1

\paragraph{Models and baselines.}
We use Qwen3-VL-2B-Instruct and Qwen3-VL-4B-Instruct as student models, with Qwen3-VL-8B-Instruct serving as the teacher~\citep{bai2025qwen3vl}. For each student size, we conduct on-policy distillation with either the original 8B teacher or an enhanced teacher obtained by training the 8B model with GRPO~\citep{shao2024deepseekmath} on ViRL39K for one epoch. We compare our method against four baselines in the main experiments: the corresponding base model, GRPO, vanilla OPD~\citep{lu2025onpolicydistillation}, and VA-OPD~\citep{liu2026vaopd}. We also include OPSD~\citep{zhao2026opsd} as a baseline in Section~\ref{sec:ablation}. Details are in Appendix~\ref{app:imple_details}.

\subsection{Main Results}
\label{sec:main-results}

Table~\ref{tab:main-results} reports results across all benchmarks for both 2B and 4B students. We compare S-OPD against the corresponding distillation baseline under the same teacher configuration. 

\paragraph{S-OPD benefits different student scales and teacher strengths.}
S-OPD improves average performance across all four student--teacher configurations. With the 8B teacher, the gains are larger for the 2B student than for the 4B student (+1.37 vs.\ +0.61 points), suggesting greater room for improvement through student-side learning at the smaller scale. Strengthening the teacher with GRPO raises the OPD baseline, yet S-OPD provides further gains of 0.92 and 0.82 points respectively. The gain decreases for the 2B student but increases for the 4B student, indicating that the benefit of student-side optimization depends on both student capacity and teacher strength.

\paragraph{Gains span reasoning domains but vary across tasks.}
S-OPD improves the average score within mathematical, logical, and general reasoning in every student--teacher configuration. The distribution of gains nevertheless differs across scales and teachers. With the 8B teacher, the 2B student improves on all eight benchmarks, including MathVista (+1.80) and ZeroBench (+2.70). With the GRPO-trained teacher, the most pronounced gain shifts to logical reasoning for the 4B student, which improves by 4.25 points on LogicVista, while its general reasoning performance changes little overall. These results suggest that student-side perceptual learning benefits diverse reasoning tasks, with the largest gains depending on the student--teacher configuration.

\paragraph{Student-side learning complements enhanced teacher supervision.}
Adding our objectives to VA-OPD improves average performance by 1.03 points for the 2B student and 0.72 points for the 4B student, with notable improvements including MathVision (+2.70) for the 2B student and MMMU-val (+1.67) for the 4B student. These gains show that improving teacher supervision leaves additional opportunities on the student side: S-OPD remains effective when combined with a method that already enhances the supervision students receive.


\newcommand{\valdiff}[2]{#1\scriptsize\,#2}
\newcommand{\gain}[1]{\textcolor{green!60!black}{(#1)}}
\newcommand{\loss}[1]{\textcolor{red!70!black}{(#1)}}
\begin{table*}[t]
\centering
\caption{Main results on eight benchmarks. We compare the distillation performance against the corresponding baseline. For OPD and S-OPD, we use either Qwen3-VL-8B-Instruct or its GRPO-trained variant as the teacher; the subscript \(G\) denotes the latter teacher setting. Avg. is computed over the eight benchmarks.}
\label{tab:main-results}
\footnotesize
\setlength{\tabcolsep}{2.3pt}
\resizebox{\textwidth}{!}{%
\begin{tabular}{lccccccccc}
\toprule
& \multicolumn{4}{c}{\textbf{Mathematical Reasoning}}
& \multicolumn{2}{c}{\textbf{Logical Reasoning}}
& \multicolumn{2}{c}{\textbf{General Reasoning}}
& \textbf{Overall} \\
\cmidrule(lr){2-5}
\cmidrule(lr){6-7}
\cmidrule(lr){8-9}
\cmidrule(lr){10-10}
\textbf{Method}
& \textbf{MathVerse}
& \textbf{MathVista}
& \textbf{MathVision}
& \textbf{WeMath}
& \textbf{LogicVista}
& \textbf{VisualPuzzles}
& \textbf{ZeroBench}
& \textbf{MMMU}
& \textbf{Avg.} \\
\midrule
\multicolumn{10}{c}{\textit{\textbf{Qwen3-VL-2B-Instruct}}} \\
\midrule
Base model
& 45.51 & 61.2 & 30.29 & 32.29
& 34.52 & 16.18 & 13.17 & 45.78 & 34.87 \\
GRPO
& 48.43 & 62.3 & 31.62 & 36.57
& 42.73 & 23.16 & 14.67 & 49.67 & 38.64 \\
OPD
& 46.41 & 65.2 & 34.35 & 36.67
& 36.24 & 12.16 & 9.28 & 48.44 & 36.09 \\
\rowcolor{myblue}
\textbf{S-OPD}
& \valdiff{47.81}{\gain{+1.40}} & \valdiff{67.0}{\gain{+1.8}}
& \valdiff{35.93}{\gain{+1.58}} & \valdiff{37.43}{\gain{+0.76}}
& \valdiff{37.65}{\gain{+1.41}} & \valdiff{12.67}{\gain{+0.51}}
& \valdiff{11.98}{\gain{+2.70}} & \valdiff{49.22}{\gain{+0.78}}
& \valdiff{37.46}{\gain{+1.37}} \\
OPD$_G$
& 50.68 & 67.6 & 35.95 & 39.24
& 39.49 & 14.89 & 12.57 & 47.33 & 38.47 \\
\rowcolor{myblue}
\textbf{S-OPD$_G$}
& \valdiff{52.76}{\gain{+2.08}} & \valdiff{68.3}{\gain{+0.7}}
& \valdiff{36.03}{\gain{+0.08}} & \valdiff{40.29}{\gain{+1.05}}
& \valdiff{39.25}{\loss{-0.24}} & \valdiff{15.62}{\gain{+0.73}}
& \valdiff{13.78}{\gain{+1.21}} & \valdiff{49.11}{\gain{+1.78}}
& \valdiff{39.39}{\gain{+0.92}} \\
\midrule
VA-OPD
& 45.48 & 63.7 & 31.77 & 33.14
& 36.91 & 10.79 & 11.38 & 47.56 & 35.09 \\
\rowcolor{myblue}
\quad \textbf{w/ S-OPD}
& \valdiff{47.51}{\gain{+2.03}} & \valdiff{64.3}{\gain{+0.6}}
& \valdiff{34.47}{\gain{+2.70}} & \valdiff{34.19}{\gain{+1.05}}
& \valdiff{38.03}{\gain{+1.12}} & \valdiff{11.56}{\gain{+0.77}}
& \valdiff{10.78}{\loss{-0.60}} & \valdiff{48.11}{\gain{+0.55}}
& \valdiff{36.12}{\gain{+1.03}} \\
\midrule
\multicolumn{10}{c}{\textit{\textbf{Qwen3-VL-4B-Instruct}}} \\
\midrule
Base model
& 42.28 & 73.8 & 49.31 & 55.43
& 54.81 & 25.43 & 15.57 & 61.56 & 47.27 \\
GRPO
& 39.11 & 73.5 & 48.39 & 52.57
& 59.73 & 41.35 & 20.36 & 62.78 & 49.72 \\
OPD
& 46.14 & 74.2 & 47.33 & 50.95
& 54.81 & 29.71 & 19.46 & 61.44 & 48.01 \\
\rowcolor{myblue}
\textbf{S-OPD}
& \valdiff{45.84}{\loss{-0.30}} & \valdiff{75.2}{\gain{+1.0}}
& \valdiff{47.57}{\gain{+0.24}} & \valdiff{52.38}{\gain{+1.43}}
& \valdiff{55.93}{\gain{+1.12}} & \valdiff{30.22}{\gain{+0.51}}
& \valdiff{20.05}{\gain{+0.59}} & \valdiff{61.78}{\gain{+0.34}}
& \valdiff{48.62}{\gain{+0.61}} \\
OPD$_G$
& 46.60 & 76.6 & 48.55 & 55.71
& 54.14 & 33.99 & 20.36 & 64.33 & 50.04 \\
\rowcolor{myblue}
\textbf{S-OPD$_G$}
& \valdiff{46.81}{\gain{+0.21}} & \valdiff{77.0}{\gain{+0.4}}
& \valdiff{48.22}{\loss{-0.33}} & \valdiff{57.43}{\gain{+1.72}}
& \valdiff{58.39}{\gain{+4.25}} & \valdiff{34.25}{\gain{+0.26}}
& \valdiff{20.05}{\loss{-0.31}} & \valdiff{64.76}{\gain{+0.43}}
& \valdiff{50.86}{\gain{+0.82}} \\
\midrule
VA-OPD
& 44.01 & 72.6 & 46.45 & 51.05
& 52.80 & 30.39 & 14.67 & 60.11 & 46.51 \\
\rowcolor{myblue}
\quad \textbf{w/ S-OPD}
& \valdiff{44.49}{\gain{+0.48}} & \valdiff{73.5}{\gain{+0.9}}
& \valdiff{46.81}{\gain{+0.36}} & \valdiff{51.52}{\gain{+0.47}}
& \valdiff{53.47}{\gain{+0.67}} & \valdiff{30.99}{\gain{+0.60}}
& \valdiff{15.27}{\gain{+0.60}} & \valdiff{61.78}{\gain{+1.67}}
& \valdiff{47.23}{\gain{+0.72}} \\
\bottomrule
\end{tabular}}
\end{table*}

\subsection{Ablation Studies}
\label{sec:ablation}
\paragraph{Component analysis.}  

\begin{table}[t]
\centering
\centering
\small
\caption{Component analysis with 2B and 4B students on Geometry3K. Avg. denotes the arithmetic average over all benchmarks including the test split of Geometry3K.}
\label{tab:component-ablation}
\setlength{\tabcolsep}{3pt}
\resizebox{\textwidth}{!}{%
\begin{tabular}{lcccccccccc}
\toprule
& \multicolumn{5}{c}{\textbf{2B Student}}
& \multicolumn{5}{c}{\textbf{4B Student}} \\
\cmidrule(lr){2-6}\cmidrule(lr){7-11}
\textbf{Method}
& \textbf{Geometry3K}
& \textbf{Mathematical}
& \textbf{Logical}
& \textbf{General}
& \textbf{Avg.}
& \textbf{Geometry3K}
& \textbf{Mathematical}
& \textbf{Logical}
& \textbf{General}
& \textbf{Avg.} \\
\midrule
OPD       & 40.93 & 45.98 & 25.83 & 30.40 & 37.48 & 46.76 & 54.38 & 42.89 & 36.76 & 47.06 \\
\quad w/ TPC       & 42.56 & 47.10 & 26.05 & 29.77 & 38.07 & \textbf{49.02} & 54.94 & 42.88 & 37.99 & 47.83 \\
\quad w/ PA       & 41.93 & 46.78 & 27.91 & \textbf{31.27} & 38.60 & 48.03 & 54.42 & 41.46 & \textbf{39.06} & 47.41 \\
\rowcolor{myblue}
\textbf{S-OPD}    & \textbf{44.26} &\textbf{47.39} & \textbf{28.02} & 30.92 & \textbf{39.08} & \textbf{49.02}& \textbf{55.03} & \textbf{43.33} & 38.57 & \textbf{48.10} \\
\bottomrule
\end{tabular}%
}
\end{table}
Table~\ref{tab:component-ablation} isolates the contributions of TPC and PA across both 2B and 4B student scales, and shows that TPC and PA contribute differently to in-domain learning and out-of-domain generalization. TPC provides stronger in-domain gains, improving Geometry3K accuracy by 1.63 and 2.26 points for the 2B and 4B students, respectively, compared with 1.00 and 1.27 points from PA. PA is particularly effective for transfer to general reasoning, achieving the highest scores at both scales, and also improves logical reasoning for the 2B student. These gains suggest that encouraging consistency under visual noise helps students generalize beyond the training domain. Combining both modules improves over OPD on Geometry3K and all three other categories at both scales, yielding average gains of 1.60 and 1.04 points. The results support their complementary roles: TPC strengthens learning from visual evidence within the training domain, while PA promotes perceptual stability that supports cross-domain transfer.

\begin{table*}[t]
\centering
\caption{Comparison of gating strategies for TPC with the remaining S-OPD configuration fixed. OPD is included as a reference. Random gating selects the same number of tokens as teacher-calibrated gating for each response. Bold values indicate the best results.}
\label{tab:gating-ablation}
\setlength{\tabcolsep}{4pt}
\resizebox{.9\textwidth}{!}{%
\begin{tabular}{lcccccc}
\toprule
\textbf{Gating strategy} & \textbf{Token selection} & \textbf{Geometry3K} & \textbf{Mathematical} & \textbf{Logical} & \textbf{General} & \textbf{Avg.} \\
\midrule
OPD (reference)
& -- & 40.93 & 45.98 & 25.83 & 30.40 & 37.48 \\
\midrule
No gating
& All & 42.56 & 46.37 & 26.55 & 30.72 & 38.06 \\
Student-based
& Adaptive & 41.43 & 46.80 & 25.44 & 30.92 & 37.93 \\
Random
& Count-matched & 40.61 & \textbf{47.47} & 26.57 & \textbf{30.95} & 38.39 \\
\rowcolor{myblue}
\textbf{Teacher-calibrated (Ours)}
& Adaptive & \textbf{44.26} & 47.39 & \textbf{28.02} & 30.92 & \textbf{39.08} \\
\bottomrule
\end{tabular}}
\end{table*}
\begin{wrapfigure}[14]{r}{0.43\columnwidth}
    \centering
    \vspace{-12pt}
\includegraphics[width=\linewidth]{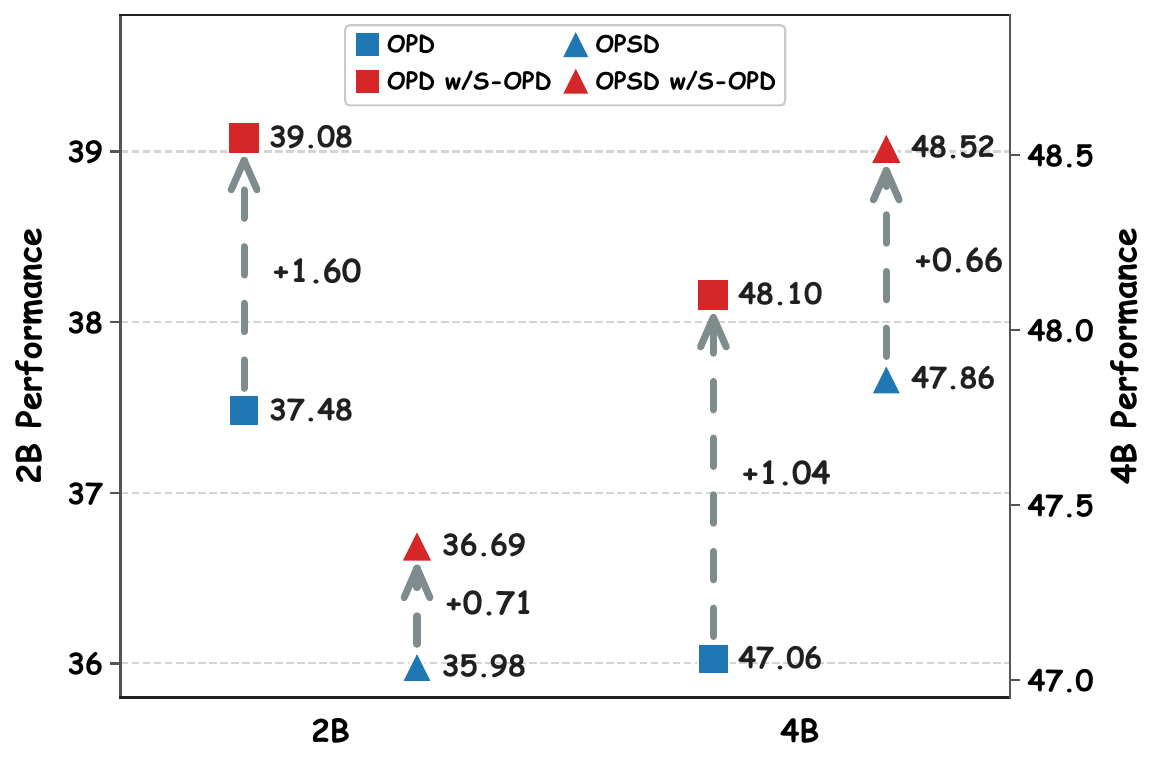}
    \caption{S-OPD gains across student scales and distillation paradigms.}
    \label{fig:scale_and_paradigm}
\end{wrapfigure}
\paragraph{Effectiveness of teacher-calibrated gating.}
We examine whether the teacher provides useful guidance on where to apply policy contrast (PC). As shown in Table~\ref{tab:gating-ablation}, teacher-calibrated gating outperforms student-based gating and no gating by 1.15 and 1.02 average points, respectively. More importantly, comparing these results with the PA-only variant in Table~\ref{tab:component-ablation} reveals that TPC is not inherently beneficial: applying PC to all tokens reduces the average score from 38.60 to 38.06. This comparison shows that teacher calibration is critical for preventing indiscriminate contrastive supervision from harming the student policy.
Student-based gating offers no improvement over applying TPC to all tokens, suggesting that the student's own visual sensitivity is insufficient to identify where policy contrast is beneficial. These results support \textit{using the teacher already available in OPD to guide the student's perceptual learning} beyond providing distillation targets. To test whether the gains simply arise from selecting fewer tokens, we randomly select the same number of tokens as the teacher gate for each response and reassign the original teacher-derived weights to these randomly selected tokens. This matched random baseline falls short of teacher-calibrated gating by 0.69 average points, indicating that the teacher identifies informative tokens for TPC.

\paragraph{Compatibility across distillation paradigms and student scales.}
Figure~\ref{fig:scale_and_paradigm} evaluates S-OPD across different student scales and teacher advantage sources trained on Geometry3K. Standard OPD obtains supervision from a larger teacher, whereas OPSD~\citep{zhao2026opsd} uses privileged information to strengthen supervision in a self-teacher setting. S-OPD improves both paradigms: under standard OPD, it increases the average performance of the 2B and 4B students by 1.60 and 1.04 points, respectively; under OPSD, the corresponding gains are 0.71 and 0.66 points. Larger gains under standard OPD suggest that student-side optimization offers more room for improvement in this setting, while the positive gains under OPSD show that it remains effective when teacher advantage comes from additional information rather than greater model capacity. Together, these results demonstrate the applicability of S-OPD across different student sizes and forms of teacher supervision, with the extent of its benefit depending on the distillation setting.

\begin{wraptable}[11]{r}{0.45\textwidth}
\centering
\vspace{-12pt}
\caption{Ablation of the TPC and PA coefficients.}
\label{tab:coefficient-ablation}
\small
\setlength{\tabcolsep}{3pt}
\renewcommand{\arraystretch}{1.1}
\resizebox{\linewidth}{!}{%
\begin{tabular}{@{}cccccc@{}}
\toprule
\textbf{Value} & \textbf{Geometry3K} &\textbf{ Mathematical} & \textbf{Logical} & \textbf{General} & \textbf{Avg.} \\
\midrule
\rowcolor{gray!15}
\multicolumn{6}{c}{$\lambda_{\text{TPC}}$} \\
\midrule
\rowcolor{myblue}
\textbf{0.005} & 44.26 & 47.39 & \textbf{28.02} & 30.92 & \textbf{39.08} \\
0.01 & \textbf{44.92} & \textbf{47.73} & 25.83 & \textbf{30.93} & 38.82 \\
0.02 & 43.05 & 47.00 & 26.64 & 30.36 & 38.34 \\
\midrule
\rowcolor{gray!15}
\multicolumn{6}{c}{$\lambda_{\text{PA}}$} \\
\midrule
0.01 & \textbf{44.92} & 46.85 & 26.55 & 30.91 & 38.58 \\
\rowcolor{myblue}
\textbf{0.02} & 44.26 & \textbf{47.39} & \textbf{28.02} & 30.92 & \textbf{39.08} \\
0.04 & 44.09 & 47.12 & 26.13 & \textbf{31.38} & 38.62 \\
\bottomrule
\end{tabular}%
}
\end{wraptable}
\paragraph{Impact of the coefficients of TPC and PA.} 
We examine whether S-OPD requires precise coefficient tuning by varying $\lambda_{\text{TPC}}$ and $\lambda_{\text{PA}}$ for students trained on Geometry3K. As shown in Table~\ref{tab:coefficient-ablation}, average performance varies by only 0.74 points for TPC and 0.50 points for PA. Although individual domains favor different coefficients, the overall performance remains stable, allowing us to obtain comparable results across a range of objective weights. We obtain the highest average score with $\lambda_{\text{TPC}}=0.005$ and $\lambda_{\text{PA}}=0.02$ on Geometry3K. We set both coefficients to $0.02$ for the main experiments on ViRL39K and the corresponding ablations are in Appendix~\ref{app:impact_of_coef}.

\newcommand{\heatcell}[2]{\cellcolor{blue!#1}#2}
\begin{wraptable}[12]{r}{0.45\textwidth}
\centering
\vspace{-12pt}
\caption{Ablation of the masking ratio and Gaussian noise level.}
\label{tab:perturbation-ablation}
\small
\setlength{\tabcolsep}{3pt}
\renewcommand{\arraystretch}{1.1}
\resizebox{\linewidth}{!}{%
\begin{tabular}{@{}cccccc@{}}
\toprule
\textbf{Value} & \textbf{Geometry3K} & \textbf{Mathematical}
& \textbf{Logical} & \textbf{General} & \textbf{Avg.} \\
\midrule
\multicolumn{6}{c}{\textbf{Masking ratio} $\rho$} \\
\midrule
0.2
& \heatcell{14}{43.43}
& \heatcell{3}{47.11}
& \heatcell{3}{24.94}
& \heatcell{25}{\textbf{32.34}}
& \heatcell{3}{38.49} \\

0.4
& \heatcell{3}{42.43}
& \heatcell{9}{47.18}
& \heatcell{15}{26.65}
& \heatcell{7}{31.20}
& \heatcell{5}{38.54} \\

\textbf{0.6}
& \heatcell{23}{44.26}
& \heatcell{25}{\textbf{47.39}}
& \heatcell{25}{\textbf{28.02}}
& \heatcell{3}{30.92}
& \heatcell{25}{\textbf{39.08}} \\

0.8
& \heatcell{25}{\textbf{44.43}}
& \heatcell{21}{47.34}
& \heatcell{11}{26.02}
& \heatcell{11}{31.42}
& \heatcell{12}{38.74} \\
\midrule
\multicolumn{6}{c}{\textbf{Gaussian noise level} $\sigma$} \\
\midrule
0.1
& \heatcell{3}{42.93}
& \heatcell{3}{46.85}
& \heatcell{8}{26.55}
& \heatcell{3}{30.91}
& \heatcell{3}{38.36} \\

\textbf{0.2}
& \heatcell{25}{\textbf{44.26}}
& \heatcell{25}{\textbf{47.39}}
& \heatcell{25}{\textbf{28.02}}
& \heatcell{3}{30.92}
& \heatcell{25}{\textbf{39.08}} \\

0.3
& \heatcell{5}{43.05}
& \heatcell{14}{47.12}
& \heatcell{3}{26.13}
& \heatcell{25}{\textbf{31.38}}
& \heatcell{8}{38.51} \\
\bottomrule
\end{tabular}%
}
\end{wraptable}
\paragraph{Impact of the masking ratio and Gaussian noise level.} 
We vary the masking ratio and Gaussian noise level to identify effective perturbation strengths for TPC and PA. In Table~\ref{tab:perturbation-ablation}, increasing $\rho$ from 0.2 to 0.6 raises the average score from 38.49 to 39.08, but further increasing it to 0.8 lowers the score to 38.74. Similarly, Gaussian noise performs best at $\sigma=0.2$, with weaker and stronger noise yielding lower average scores. We thus obtain the best overall results with substantial masking for TPC and moderate noise for PA. The preferred strength varies by domain, suggesting that different tasks may place distinct demands on visual perception. We choose $\rho=0.6$ and $\sigma=0.2$ for both Geometry3K ablations and ViRL39K main experiments.

\paragraph{Computational overhead analysis.} 
S-OPD improves reasoning performance without adding trainable parameters, extra autoregressive rollouts, or inference overhead. Its additional computation is confined to training, where teacher gating and the auxiliary student objectives evaluate perturbed image views using the same generated responses. Under identical hardware configurations, these operations increase training time per step by 28.6\% for the 2B student on ViRL39K and 16.0\% for the 4B student on Geometry3K, at a moderate cost while preserving deployment efficiency. Appendix~\ref{app:computational_cost} provides detailed measurements.

\section{Conclusion}
\label{sec:conclusion}
In this paper, we identify student visual perception as a persistent bottleneck in multimodal on-policy distillation, even when teacher supervision is strengthened. Motivated by this, we propose S-OPD to explicitly improve student perceptual sensitivity and robustness within the token-level distillation process. Teacher-calibrated Policy Contrast encourages sensitivity to visual evidence removal at teacher-selected tokens, while Policy Agreement promotes consistent predictions under visual noise. Experiments on eight benchmarks demonstrate consistent average improvements across student scales and teacher strengths, with further gains when integrated into existing methods that enhance teacher supervision. These benefits require no additional annotations, model parameters, or inference-time operations, with only moderate training overhead. Our findings establish student-side learning as a complementary direction for advancing multimodal OPD.



\bibliography{iclr2027_conference}
\bibliographystyle{iclr2027_conference}

\clearpage

\appendix

\section{Implementation Details}
\label{app:imple_details}
\subsection{Training Details}
\label{app:train_details}

All the experiments are implemented on top of the \texttt{verl} framework~\citep{sheng2025hybridflow} and run on NVIDIA A100 80GB GPUs. We train the student model with PyTorch Fully Sharded Data Parallel (FSDP)~\citep{zhao2023pytorch} and use vLLM~\citep{kwon2023vllm} for on-policy rollout generation and frozen-teacher forward passes. Table~\ref{tab:training-config} summarizes the key training configurations for the main ViRL39K experiments and the Geometry3K ablations.

\begin{table}[ht]
\centering
\caption{Key training configurations for the ViRL39K experiments and the Geometry3K ablations.}
\label{tab:training-config}
\small
\setlength{\tabcolsep}{6pt}
\begin{tabular}{lccc}
\toprule
Setting & ViRL39K Distillation & ViRL39K GRPO & Geometry3K Distillation \\
\midrule
Training data & ViRL39K & ViRL39K & Geometry3K train\\
Max. prompt length & 4,096 & 4,096 & 1,024 \\
Max. response length & 4,096 & 4096 & 2,048\\
Prompt batch size & 192 & 192 & 128 \\
Learning rate & $1\times10^{-6}$ & $1\times10^{-6}$ & $1\times10^{-6}$ \\
Rollouts per prompt ($n$) & 1 & 5 & 1\\
Training epochs & 1 & 1 & 20\\
$\lambda_\text{TPC}$ & 0.02 & -- & 0.005 \\
$\lambda_\text{PA}$ & 0.02 & -- & 0.02 \\
\bottomrule
\end{tabular}
\end{table}

\begin{figure}[H]
    \centering
    \includegraphics[width=.8\linewidth]{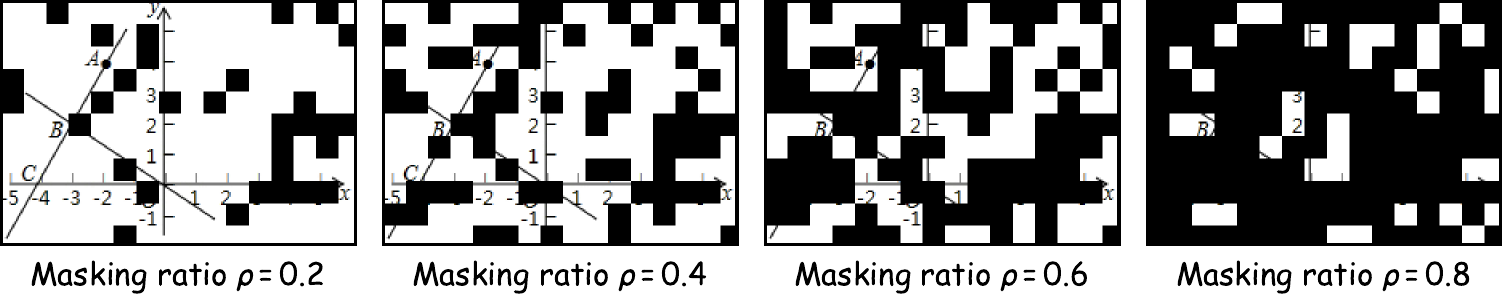}
    \caption{Masked images used for TPC at different masking ratios $\rho$. Randomly selected image patches are replaced with black pixels, removing increasing amounts of visual evidence as $\rho$ increases.}
    \label{fig:mask_ratio_visualization}
\end{figure}

\paragraph{Details of masking.}
We construct the masked image in RGB pixel space before applying the model-specific image processor. Specifically, we partition the original-resolution image into non-overlapping $16\times16$ pixel blocks, including partial blocks along the image boundaries, and independently sample a mask indicator $m_{ij}\sim\operatorname{Bernoulli}(0.6)$ for each block. If $m_{ij}=1$, all pixels in that block are replaced with black pixels, i.e., RGB $(0,0,0)$. Thus, $0.6$ denotes the expected masking ratio rather than an exact per-image proportion. We modify only the pixel values and preserve the image dimensions; no image patches or visual tokens are removed. Consequently, the original and masked images have the same visual-token sequence length and two-dimensional positional structure after preprocessing. Figure~\ref{fig:mask_ratio_visualization} visualizes images under different masking ratios.

\begin{figure}[t]
    \centering
    \includegraphics[width=.6\linewidth]{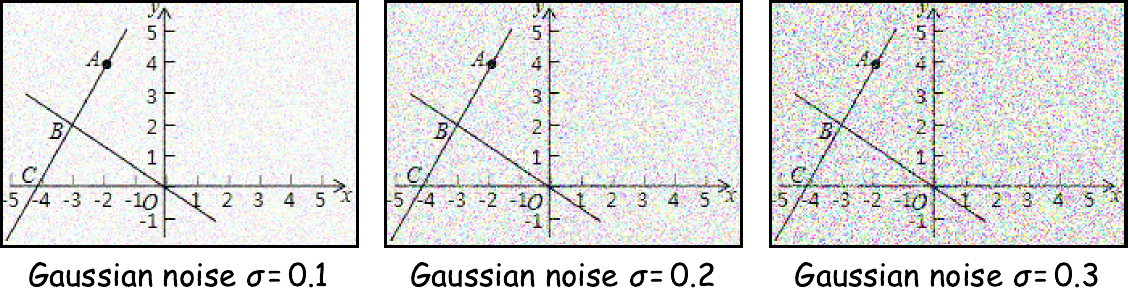}
    \caption{Noisy images used for PA at different Gaussian noise levels $\sigma$. Independent zero-mean Gaussian noise is added to each RGB channel in the $[0,1]$ pixel range, followed by clipping. Larger $\sigma$ produces stronger perturbations while retaining the overall image structure.}
    \label{fig:gaus_visualization}
\end{figure}

\paragraph{Details of Gaussian perturbation.}
We construct the noisy image by adding Gaussian noise before applying the model-specific image processor. Given an unsigned 8-bit RGB image $v$, we first rescale its pixel values from $[0,255]$ to $[0,1]$ and independently sample
$\epsilon_{hwc}\sim\mathcal{N}(0,\sigma^2)$ for every spatial location and color channel, with $\sigma=0.2$. The perturbed image is computed as
\begin{equation}
v_{\mathrm n}
=
\operatorname{round}\!\left[
255\cdot
\operatorname{clip}\!\left(
\frac{v}{255}+\epsilon,\,0,\,1
\right)
\right],
\end{equation}
and is converted back to an unsigned 8-bit RGB image before standard visual preprocessing. Therefore, the reported noise scale is defined with respect to the $[0,1]$ pixel range; equivalently, $\sigma=0.2$ corresponds to approximately $51$ intensity levels on the $[0,255]$ scale. Figure~\ref{fig:gaus_visualization} visualizes images under different Gaussian perturbation noise.

\subsection{Evaluation Details}
\label{app:eval_details}
All the checkpoints are evaluated on \path{MathVerse_MINI}, \path{MathVista_MINI}, \path{MathVision}, \path{WeMath}, \path{LogicVista}, \path{VisualPuzzles}, \path{ZeroBench_sub}, \path{MMMU_DEV_VAL} with greedy decoding using the default VLMEvalKit~\citep{duan2024vlmevalkit} dataset definitions and a common Qwen3-VL wrapper in the main experiments. We use the shared system prompt \texttt{You are a helpful assistant.} and dataset-specific user prompts listed in Table~\ref{tab:evaluation-prompts}. 

\begin{table}[t]
\centering
\caption{Generation user prompts used for each evaluation dataset. Braced terms
denote fields from each VLMEvalKit example, and \texttt{[Options]} denotes
the formatted list of answer choices.}
\label{tab:evaluation-prompts}
\footnotesize
\renewcommand{\arraystretch}{1.15}
\begin{tabularx}{\linewidth}{@{}p{0.22\linewidth}>{\ttfamily\raggedright\arraybackslash}X@{}}
\toprule
Dataset & \normalfont User prompt template \\
\midrule
\texttt{MathVerse\_MINI}
& \{question\} \\
\texttt{MathVista\_MINI}
& \{question\} \\
\texttt{MathVision}
& \{question\} \\
\texttt{WeMath}
& Question: \{question\} \newline
  Options: [Options] \\
\texttt{LogicVista}
& \{question\} \\
\texttt{VisualPuzzles}
& \{question\} \newline
  Options: [Options] \newline
  Solve the multiple-choice question and then answer with the option letter
  from the given choices. The last line of your response should be of the
  following format: 'Answer: \$LETTER' (without quotes), where LETTER is one
  of the options. Think step by step before answering. \\
\texttt{ZEROBench\_sub}
& \{question\} \newline
  Let's think step by step and give the final answer in curly braces, like
  this: \{final answer\}. \\
\texttt{MMMU\_DEV\_VAL}
& Question: \{question\} \newline
  Options: [Options] \newline
  Please select the correct answer from the options above. \\
\bottomrule
\end{tabularx}
\end{table}

Rule-based benchmarks use their official exact-match or multiple-choice scorers. Benchmarks requiring answer extraction or semantic grading use a locally served Qwen3-30B-A3B. MathVerse is averaged over its Vision
Intensive, Vision Only, Text Dominant, Text Lite, and Vision Dominant subsets; the text-only subset is excluded. We report MathVista on its 1,000-example mini split, WeMath using Core (Strict), and MMMU on the validation split.

\subsection{Oracle Visual-Fact Intervention Details}
\label{app:probe_details}

\paragraph{Oracle-fact construction.}
We use oracle visual facts as a test-time intervention to make the geometric information encoded in each diagram explicit. For every Geometry3K example, we deterministically convert \texttt{diagram\_logic\_form} from the official \texttt{logic\_form.json} annotations into natural-language statements describing geometric entities, incidence relations, lengths, angles, parallelism, perpendicularity, equality, congruence, and other geometric properties. The renderer has no access to the answer, solution, program, theorem sequence, or text-based logic forms. It also rejects \texttt{Find} and \texttt{UseTheorem} expressions, including nested occurrences. We compare two paired inputs that share the same image, question, answer choices, system prompt, and output instruction. The oracle condition differs only by the insertion of the rendered statements under \texttt{Diagram observations:} in the same user message. Thus, the intervention augments the diagram information available to the model without changing its parameters or removing the original image.

\paragraph{Evaluation protocol.}
We evaluate every checkpoint on the same official 601-example test split of Geometry3K. Inference uses greedy decoding (temperature 0) and a maximum generation length of 4096 tokens. We report answer-content accuracy after mapping predicted option letters to their corresponding choice values and normalizing common \LaTeX{} and numeric variants. The \emph{Accuracy Gap} is the paired difference between accuracy with oracle visual facts and image-only accuracy.

\begin{figure}[t]
\centering
\includegraphics[width=.6\linewidth]{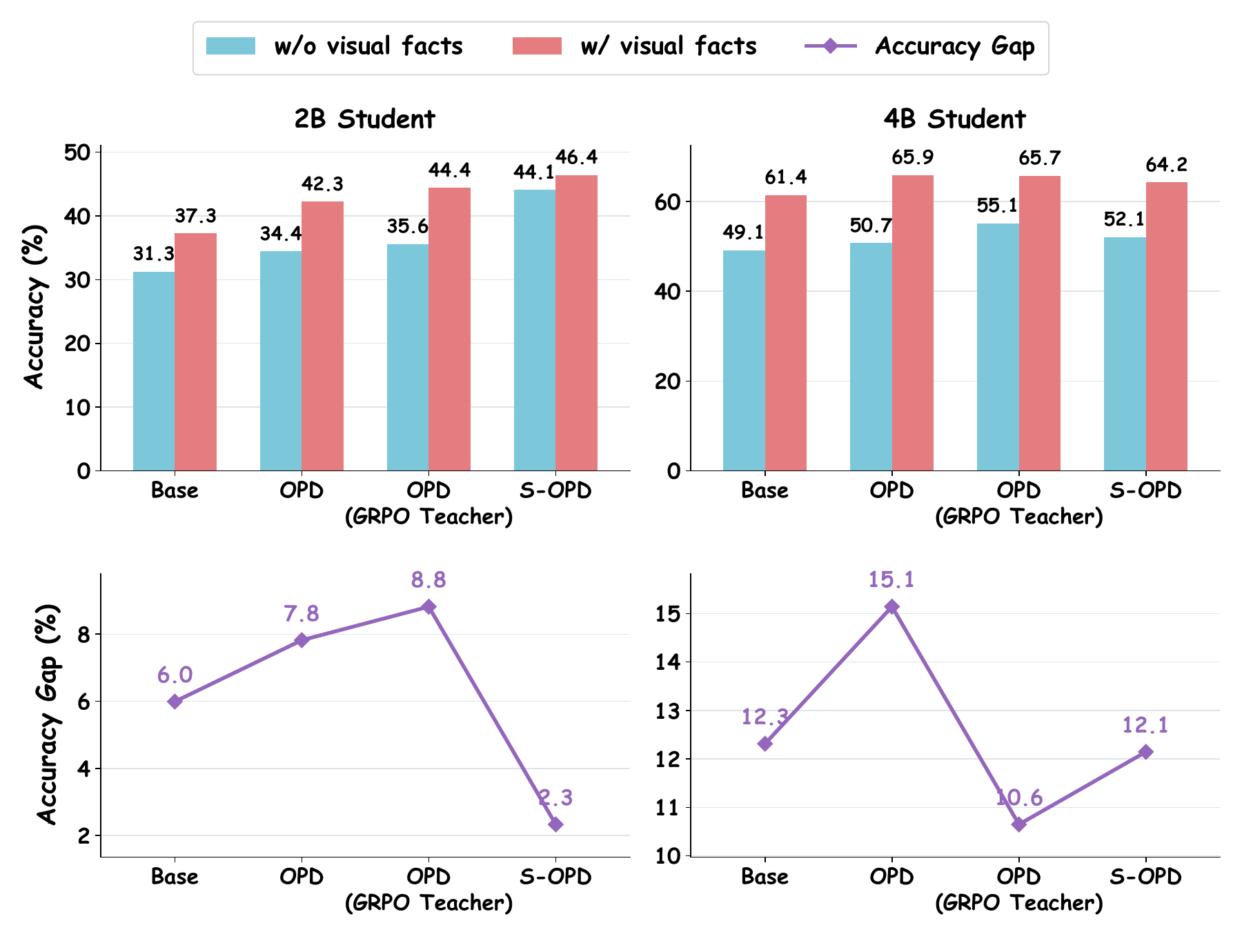}
\caption{\textbf{Oracle visual-fact intervention on Geometry3K.}
Top panels report accuracy with the original image alone (\emph{w/o visual facts}) and with additional oracle visual facts derived from diagram annotations (\emph{w/ visual facts}); bottom panels show the corresponding accuracy gap in percentage points.
Compared with standard OPD, S-OPD improves accuracy without visual facts and reduces the accuracy gap from 7.8 to 2.3 points for the 2B student and from 15.1 to 12.1 points for the 4B student.}
\label{fig:oracle_visual_facts_wSOPD}
\end{figure}

\paragraph{Results.}
We revisit the oracle visual-fact intervention from our motivational study to examine whether S-OPD alleviates the student-side perceptual bottleneck that motivated our method. As shown in Figure~\ref{fig:oracle_visual_facts_wSOPD}, compared with standard OPD, S-OPD improves image-only accuracy and reduces the Accuracy Gap at both model scales. For the 2B student, image-only accuracy increases from 34.4\% to 44.1\%, while accuracy with oracle visual facts improves from 42.3\% to 46.4\%, narrowing the gap from 7.8 to 2.3 percentage points. S-OPD thus achieves the highest accuracy in both conditions and the smallest gap among the compared methods at this scale. This result directly complements our motivational finding: whereas stronger teacher-side supervision alone leaves a substantial gap, explicitly strengthening student perception helps the student recover more task-relevant evidence from the diagram itself. For the 4B student, S-OPD improves image-only accuracy from 50.7\% to 52.1\% and reduces the gap from 15.1 to 12.1 points. Here, the evidence is more qualified, as oracle-fact accuracy also decreases from 65.9\% to 64.2\%, and OPD with the GRPO teacher retains the smallest gap. Overall, these results, particularly the strong improvement for the 2B student, support our central motivation that student-side perceptual learning is an important complement to teacher-side supervision in multimodal OPD.

\paragraph{Audit.}
Our data audit finds no unsupported expressions, parse failures, missing images, missing answers, or use of forbidden annotation sources in the rendered facts.

\section{Ablation Details}
\label{app:ablation_details}

\subsection{Visual Perception Beyond Reasoning Benchmarks}
To further examine whether S-OPD improves student visual perception, we evaluate the distilled students on three additional benchmarks, V$^\ast$Bench~\citep{wu2024vbench}, ZoomBench~\citep{wei2026zoombench} and HallusionBench~\citep{guan2024hallusionbench}. Specifically, V$^\ast$Bench and ZoomBench assess fine-grained visual understanding, while HallusionBench explicitly diagnoses failures arising from visual illusions and language hallucinations. As shown in Table~\ref{tab:other_bench_results}, S-OPD improves or matches OPD in 10 out of the 12 model--data configurations.

The improvements are consistent across different student capacities and training distributions. On V$^\ast$Bench, S-OPD improves the 2B student by 1.05 points when trained on either Geometry3K or ViRL39K, and improves the 4B student trained on ViRL39K by 0.63 points. On ZoomBench, it yields gains of 0.95 and 0.71 points for the 2B and 4B students trained on Geometry3K, respectively, while improving the ViRL39K-trained 4B student by 0.47 points. Importantly, S-OPD also consistently improves or preserves performance on HallusionBench across all four settings, with gains of up to 0.53 points. Although a few configurations exhibit small fluctuations, the overall trend across model sizes, training datasets, and diagnostic tasks indicates that the benefits of S-OPD extend beyond the in-domain reasoning benchmarks. In particular, the gains on fine-grained perception and hallucination-oriented evaluations provide additional evidence that S-OPD encourages students to ground their predictions more effectively in visual evidence.

\begin{table*}[t]
    \centering
    \caption{
        Comparison of OPD and S-OPD across different student model sizes
        and training datasets. All models are evaluated on three
        out-of-domain visual benchmarks.
    }
    \label{tab:other_bench_results}

    \footnotesize
    \setlength{\tabcolsep}{4pt}
    \renewcommand{\arraystretch}{1}
\begin{tabular}{@{}lcccccccc@{}}
        \toprule
        \multirow{3}{*}{\textbf{Benchmark}}
        & \multicolumn{4}{c}{\textbf{Trained on Geometry3K}}
        & \multicolumn{4}{c}{\textbf{Trained on ViRL39K}} \\
        \cmidrule(lr){2-5}
        \cmidrule(lr){6-9}

        & \multicolumn{2}{c}{\textbf{2B}}
        & \multicolumn{2}{c}{\textbf{4B}}
        & \multicolumn{2}{c}{\textbf{2B}}
        & \multicolumn{2}{c}{\textbf{4B}} \\
        \cmidrule(lr){2-3}
        \cmidrule(lr){4-5}
        \cmidrule(lr){6-7}
        \cmidrule(lr){8-9}

        & OPD & S-OPD
        & OPD & S-OPD
        & OPD & S-OPD
        & OPD & S-OPD \\
        \midrule

        V$^{*}$Bench
        & 73.30 & \textbf{74.35}
        & \textbf{74.87} & 74.35
        & 74.87 & \textbf{75.92}
        & 62.67 & \textbf{63.30} \\

        ZoomBench
        & 41.42 & \textbf{42.37}
        & 43.55 & \textbf{44.26}
        & \textbf{41.30} & 40.95
        & 43.67 & \textbf{44.14} \\

        HallusionBench
        & 55.73 & \textbf{56.26}
        & 64.77 & \textbf{64.98}
        & 52.58 & \textbf{53.00}
        & \textbf{74.35} & \textbf{74.35} \\

        \bottomrule
    \end{tabular}
\end{table*}

\subsection{Main Component Analysis}
\label{app:main-component-analysis}
Table~\ref{tab:main-component-ablation} examines TPC and PA across both student scales. TPC improves in-domain accuracy and all four mathematical reasoning benchmarks for both students, consistent with its role in linking predictions to supporting visual evidence through teacher-calibrated policy contrast. PA yields more varied gains, with notable improvements on LogicVista and ZeroBench for the 2B student and MMMU for the 4B student. This pattern suggests that agreement under mild visual perturbations can support generalization beyond the geometric training domain, although its benefits depend on the task and student scale.

Combining both objectives achieves the highest overall average at both scales, improving over vanilla OPD by 1.60 and 1.04 points for the 2B and 4B students, respectively. Gains extend to both in-domain accuracy and out-of-domain average performance. While the combination does not preserve every component's best individual score, its stronger aggregate results support the complementary roles of sensitivity to evidence removal and stability under mild visual variation in student learning.
\begin{table*}[t]
\centering
\caption{Detailed component analysis results on Geometry3K. The Geometry3K test split is used for in-domain performance evaluation. Bold and underlined numbers indicate the best and second-best results within each student size, respectively.}
\label{tab:main-component-ablation}
\setlength{\tabcolsep}{3.5pt}
\resizebox{\textwidth}{!}{%
\begin{tabular}{lcccccccccc}
\toprule
& \textbf{In-domain}
& \multicolumn{4}{c}{\textbf{Mathematical Reasoning}}
& \multicolumn{2}{c}{\textbf{Logical Reasoning}}
& \multicolumn{2}{c}{\textbf{General Reasoning}}
& \textbf{Overall} \\
\cmidrule(lr){2-2}\cmidrule(lr){3-6}\cmidrule(lr){7-8}
\cmidrule(lr){9-10}\cmidrule(lr){11-11}
\textbf{Method} & \textbf{Geo3K-Test}
& \textbf{MVerse} & \textbf{MVista} & \textbf{MVision} & \textbf{WeMath}
& \textbf{LogicVista} & \textbf{VisPuzzles}
& \textbf{ZeroBench} & \textbf{MMMU-val} & \textbf{Avg.} \\
\midrule
\multicolumn{11}{c}{\textit{\textbf{Qwen3-VL-2B-Instruct}}} \\
\midrule
Vanilla OPD
& 40.93 & 49.28 & 63.3 & 34.96 & 36.38
& 39.83 & 11.82 & 11.68 & \textbf{49.11} & 37.48 \\
\quad w/ TPC 
& \underline{42.56} & 49.84 & \underline{65.2} & \underline{35.07}
& \textbf{38.29} & 40.71 & 11.39 & 11.98 & 47.56 & 38.07 \\
\quad w/ PA
& 41.93 & \underline{50.69} & \textbf{65.3} & 34.36 & 36.76
& \textbf{43.14} & \underline{12.67} & \textbf{14.97}
& 47.56 & \underline{38.60} \\
\rowcolor{myblue}
\textbf{S-OPD}
& \textbf{44.26} & \textbf{50.76} & \textbf{65.3}
& \textbf{35.43} & \underline{38.05}
& \underline{42.51} & \textbf{13.53} & \underline{13.17}
& \underline{48.67} & \textbf{39.08} \\
\midrule
\multicolumn{11}{c}{\textit{\textbf{Qwen3-VL-4B-Instruct}}} \\
\midrule 
Vanilla OPD
& 46.76 & 44.24 & \underline{74.1} & 46.68 & \underline{52.48}
& \underline{55.81} & \underline{29.97} & 14.07 & 59.44 & 47.06 \\
\quad w/ TPC
& \textbf{49.02} & \underline{45.14} & \textbf{74.4} & 47.07
& \textbf{53.14} & 55.70 & \textbf{30.05} & \underline{15.87}
& 60.11 & \underline{47.83} \\ 
\quad w/ PA
& \underline{48.03} & 43.83 & \underline{74.1} & \textbf{47.63}
& 52.10 & 55.26 & 27.65 & 15.57 & \textbf{62.56} & 47.41 \\
\rowcolor{myblue}
\textbf{S-OPD}
& \textbf{49.02} & \textbf{45.38} & \textbf{74.4}
& \underline{47.20} & \textbf{53.14}
& \textbf{56.60} & \textbf{30.05} & \textbf{16.47}
& \underline{60.67} & \textbf{48.10} \\
\bottomrule
\end{tabular}}
\end{table*}

\subsection{Empirical Sparsity of the Teacher Gate}
\label{app:sparsity}

We examine whether the zero-margin criterion ($\tau=0$) makes the teacher gate overly dense. In a representative S2B--T8B training run with 60\% random patch masking, we analyze 3,840 logged rollouts containing 8,649,621 valid response tokens. The gate retains 4,054,346 tokens, corresponding to a token-weighted activation rate of 46.87\%. When each rollout is weighted equally, the mean and median activation rates are 48.29\% and 48.39\%, respectively, with an interquartile range of 43.50--53.48\% and a 5th--95th percentile range of 32.36--65.64\%. These results show that masking does not induce a uniformly positive teacher likelihood drop: even with $\tau=0$, more than half of the response tokens are filtered out on average.

\subsection{Impact of the Coefficients of TPC and PA}
\label{app:impact_of_coef}
\begin{table}[t]
\centering
\caption{Ablation of the TPC and PA coefficients on ViRL39K.}
\label{tab:virl39k_coefficient_ablation}
\footnotesize
\renewcommand{\arraystretch}{0.85}
\begin{tabular*}{0.5\textwidth}{
    @{\extracolsep{\fill}} ccccc @{}
}
\toprule
\textbf{Value} & \textbf{Mathematical} & \textbf{Logical} & \textbf{General} & \textbf{Avg.} \\
\midrule
\rowcolor{gray!15}
\multicolumn{5}{c}{$\lambda_{\text{TPC}}$} \\
\midrule
0.005 & \textbf{47.07} & 25.09 & 29.28 & 37.13 \\
0.01 & 46.20 & 25.05 & 28.71 & 36.54 \\
\textbf{0.02} & 47.04 & \textbf{25.16} & \textbf{30.60} & \textbf{37.46} \\
\midrule
\rowcolor{gray!15}
\multicolumn{5}{c}{$\lambda_{\text{PA}}$} \\
\midrule
0.01 & 46.41 & 24.37 & 28.57 & 36.44 \\
\textbf{0.02} & 47.04 & \textbf{25.16} & \textbf{30.60} & \textbf{37.46} \\
0.04 & \textbf{47.15} & 24.84 & 30.44 & 37.39 \\
\bottomrule
\end{tabular*}
\end{table}
We extend the coefficient ablations on Geometry3K in the main text to ViRL39K. As shown in Table~\ref{tab:virl39k_coefficient_ablation}, average performance varies by 0.92 points across the tested TPC coefficients and 1.02 points across the PA coefficients, with all settings outperforming the OPD baseline of 36.09. Both coefficients achieve the highest average score at 0.02, while increasing $\lambda_{\text{PA}}$ to 0.04 yields nearly identical performance (37.39 vs.\ 37.46). Together with the Geometry3K results, these findings show that S-OPD remains effective across a range of coefficient choices on both training datasets. We set $\lambda_{\text{TPC}}=\lambda_{\text{PA}}=0.02$ by default in the main experiments on ViRL39K, as this configuration achieves the highest average score.

\subsection{Token-level Log-probability Analysis}
\begin{figure}
    \centering
    \includegraphics[width=\linewidth]{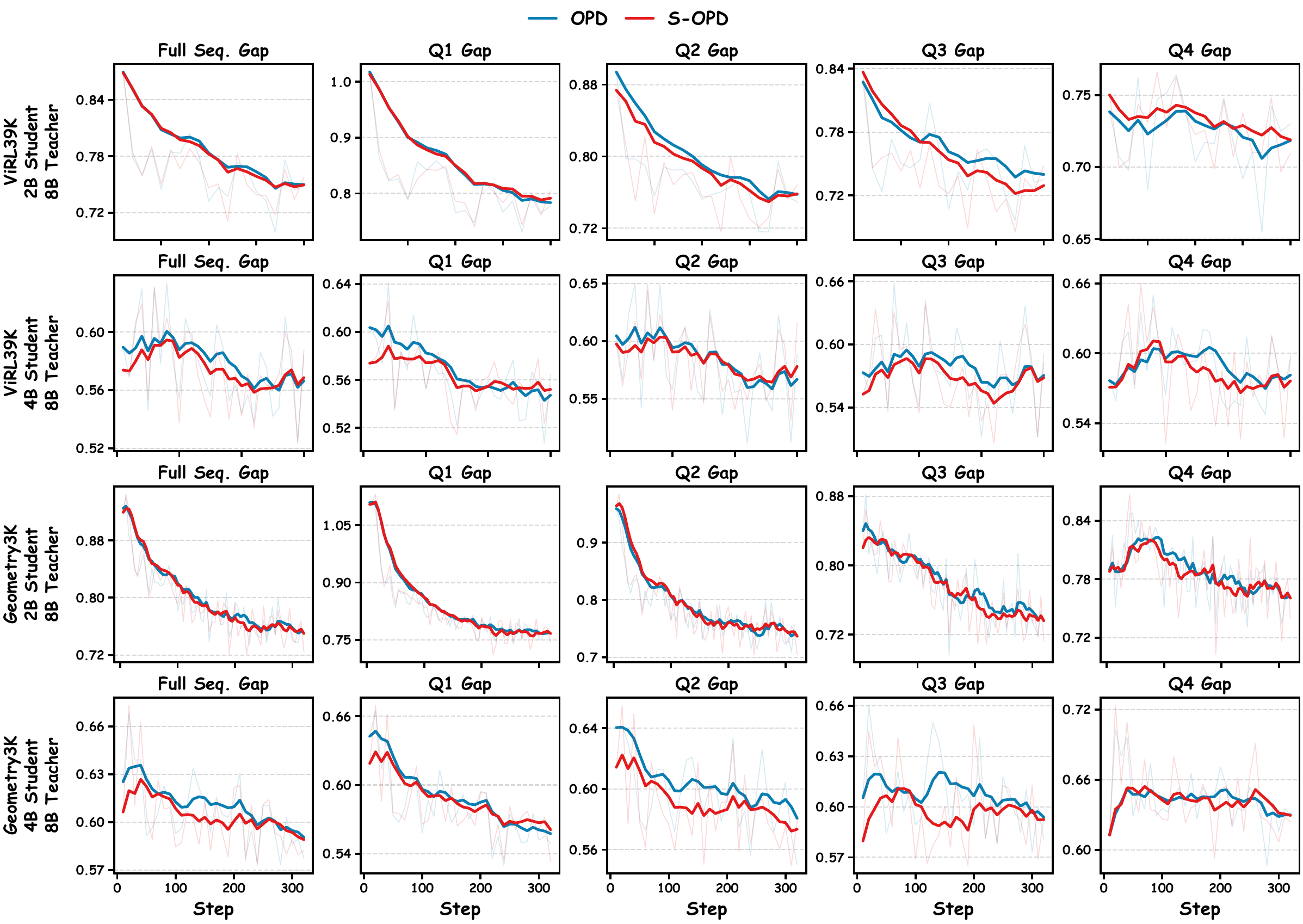}
    \caption{Token-level log-probability analysis of 2B and 4B students on ViRL39K and Geometry3K during training. We report the teacher–student token-level log-probability gap under different teacher–student configurations.}
    \label{fig:token}
\end{figure}
\paragraph{Full-rollout log-probability analysis.}
Figure~\ref{fig:token} compares the teacher--student token-level log-probability gap over complete rollouts on ViRL39K and Geometry3K. S-OPD exhibits similar training dynamics to OPD and reaches comparable final gaps at both student scales, with smaller gaps during parts of training for the 4B student, particularly on Geometry3K. Since this metric averages over initial planning, intermediate reasoning, and final-answer generation, it can obscure changes within individual stages. Together with the accuracy gains and the reductions in middle-segment gaps analyzed below, these results suggest that S-OPD preserves overall teacher alignment while improving learning within the reasoning process, consistent with its goal of strengthening the student's use of visual evidence.

\paragraph{Segment-level log-probability analysis.} 
We divide each rollout into four equal-length segments, Q1--Q4, to examine alignment throughout the response. A typical solution roughly progresses from establishing an approach, through inspecting visual evidence and deriving intermediate results, to producing the final answer, although these stages need not align exactly with segment boundaries. Under this interpretation, improvements in the middle segments are particularly relevant to learning how to connect visual evidence with reasoning. S-OPD maintains a smaller Q2 gap through much of training for the 4B student on Geometry3K and reduces the Q3 gap during later training for the 2B student on ViRL39K. Both 4B settings also show smaller Q3 gaps during portions of training. This pattern suggests that S-OPD benefits intermediate solution construction rather than concentrating its effect on final-answer prediction. Together with the comparable full-rollout gaps and improved accuracy, these observations are consistent with the intended mechanism: visual contrast and alignment help students use teacher supervision more effectively within the reasoning process.

\subsection{Computational Overhead Analysis}
\label{app:computational_cost}
S-OPD introduces additional training computation through teacher evaluation of masked images for token gating and student evaluation of masked and noisy images for TPC and PA. All branches reuse the same student-generated response, requiring no additional autoregressive rollouts, trainable parameters, or annotations. Table~\ref{tab:computational_overhead} quantifies the resulting overhead: average time per step increases from 189.44 to 243.56 seconds for the 2B student on ViRL39K and from 92.26 to 107.06 seconds for the 4B student on Geometry3K, corresponding to moderate increases of 28.6\% and 16.0\%, respectively. Average response lengths remain nearly unchanged in both settings. These auxiliary computations are used only during training; inference retains the original student architecture and decoding procedure, with no additional computational overhead.
\begin{table}[t]
\centering
\small
\setlength{\tabcolsep}{6pt}
\renewcommand{\arraystretch}{1}
\caption{Computational overhead of S-OPD. Both methods use four NVIDIA A100 80GB GPUs, two for the student and two for the teacher. Training time excludes initialization, checkpoint saving, and validation.
Overhead is relative to OPD in time per step.}
\label{tab:computational_overhead}
\begin{tabular}{@{}lrrrr@{}}
\toprule
& \multicolumn{2}{c}{\textbf{ViRL39K, 2B}}
& \multicolumn{2}{c}{\textbf{Geometry3K, 4B}} \\
\cmidrule(lr){2-3}\cmidrule(l){4-5}
\textbf{Metric} & \textbf{OPD }& \textbf{S-OPD} & \textbf{OPD} & \textbf{S-OPD} \\
\midrule
Tokens/response & 2235.9 & 2233.2 & 802.4 & 800.3 \\
Time/step (s)   & 189.44 & 243.56 & 92.26 & 107.06 \\
Gen. (ms/token) & 0.233  & 0.252  & 0.315 & 0.333 \\
\midrule
Overhead        & -- & +28.6\% & -- & +16.0\% \\
\bottomrule
\end{tabular}
\end{table}

\section{Qualitative Case Studies}
\subsection{Sample-level Rollout Analysis}
\label{app:sample_level_response}
To complement the aggregate benchmark results, we examine three sample-level cases from MathVerse, LogicVista, and MMMU-val. These examples are selected using two criteria: the OPD baseline produces an incorrect answer, whereas S-OPD produces the correct answer; and solving the problem requires information that is available only from the image. The cases cover complementary forms of visual reasoning, including reading geometric dimensions, propagating motion through spatial relations, and recognizing fine-grained visual patterns. Rather than serving as additional quantitative evidence, these examples illustrate how the two models differ in their use of visual evidence along the reasoning process.
\begin{figure}[h]
    \centering
    \includegraphics[width=\linewidth]{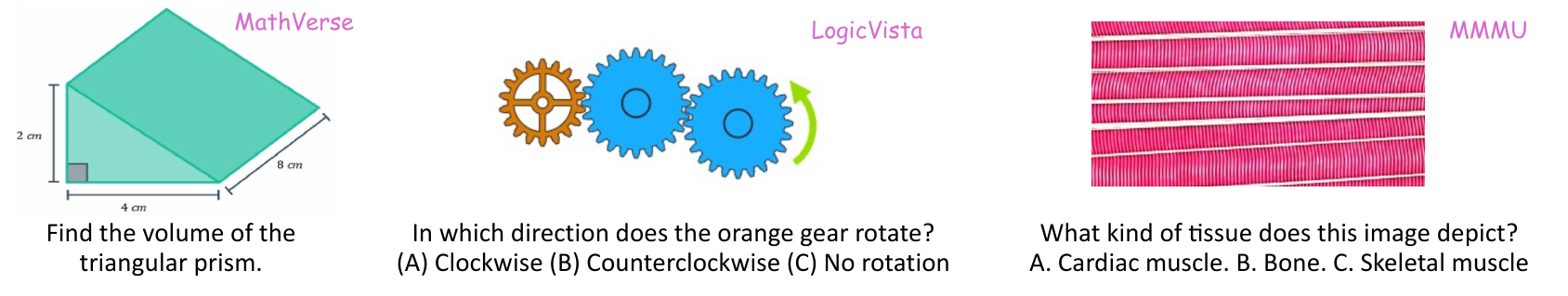}
    \caption{Qualitative examples from MathVerse, LogicVista, and MMMU-val where 2B S-OPD correctly grounds its reasoning in visual evidence, while 2B OPD produces incorrect answers.}
    \label{fig:my_label}
\end{figure}

\subsubsection{MathVerse: Triangular Prism Volume Calculation}
\begin{tcolorbox}[enhanced, breakable, frame hidden,
    borderline west={2.5pt}{0pt}{red!70!black},
    colback=red!3, left=5pt, right=3pt, top=3pt, bottom=3pt,
    before skip=5pt, after skip=5pt]
{\bfseries\small OPD baseline {\normalfont(incorrect, answer is 8 cm$^3$)}}
\smallskip\footnotesize

Official extracted answer: 8.

Although the response identified 2 cm, 4 cm, and 8 cm, it repeatedly confused the sides of the triangle with the lengths of the prism, and the final output was truncated. The original response was 14k characters long.
\end{tcolorbox}

\begin{tcolorbox}[enhanced, breakable, frame hidden,
    borderline west={2.5pt}{0pt}{blue!70!black},
    colback=blue!3, left=5pt, right=3pt, top=3pt, bottom=3pt,
    before skip=5pt, after skip=5pt]
{\bfseries\small S-OPD (Ours) {\normalfont(correct, answer is 32 cm$^3$)}}

\smallskip\footnotesize
Looking at the image:

\begin{itemize}
    \item The triangle has a base of $4$ cm and a height of $2$ cm.
    \item The side of the prism that is perpendicular to the base is $8$ cm.
\end{itemize}
\[
\text{Base Area}
=
\frac{1}{2}\times 4\times 2
=
4\text{ cm}^2
\]
\[
V=4\times 8=32\text{ cm}^3
\]

So, the volume of the triangular prism is $32$ cm$^3$.
\end{tcolorbox}

\paragraph{Analysis.}
This case tests whether the model can assign the three visible measurements to the correct geometric quantities. The OPD baseline successfully reads the values $2$, $4$, and $8$ from the image, indicating that its failure is not caused by completely missing the relevant visual content. Instead, it repeatedly confuses the dimensions of the triangular base with the length of the prism. This role-assignment error leads to an incorrect extracted answer of $8~\mathrm{cm}^3$ and a long, self-correcting response that ultimately terminates without a consistent derivation.

S-OPD uses the same visual measurements more coherently. It identifies $4$ cm and $2$ cm as the base and height of the right-triangular face, respectively, and interprets $8$ cm as the length of the prism. It therefore computes the triangular area as $\frac{1}{2}\times4\times2=4~\mathrm{cm}^2$ and the volume as $4\times8=32~\mathrm{cm}^3$. The contrast between the responses suggests that, for this example, \textbf{S-OPD improves the integration and semantic assignment of perceived quantities rather than merely extracting more numbers from the image}.

\subsubsection{LogicVista: Gear Rotation Direction Reasoning}
\begin{tcolorbox}[enhanced, breakable, frame hidden,
    borderline west={2.5pt}{0pt}{red!70!black},
    colback=red!3, left=5pt, right=3pt, top=3pt, bottom=3pt,
    before skip=5pt, after skip=5pt]
{\bfseries\small OPD baseline {\normalfont(incorrect, answer is A)}}
\smallskip\footnotesize

The rightmost blue gear is rotating counterclockwise, so the middle blue gear must rotate clockwise, and the orange gear must rotate clockwise as well.

Answer: (A) Clockwise
\end{tcolorbox}
\begin{tcolorbox}[enhanced, breakable, frame hidden,
    borderline west={2.5pt}{0pt}{blue!70!black},
    colback=blue!3, left=5pt, right=3pt, top=3pt, bottom=3pt,
    before skip=5pt, after skip=5pt]
{\bfseries\small S-OPD (Ours) {\normalfont(correct, answer is B)}}

\smallskip\footnotesize
- The rightmost blue gear is rotating counterclockwise, as indicated by the green arrow.

- Since it is meshed with the middle blue gear, the middle blue gear must rotate clockwise.

- The middle blue gear is also meshed with the orange gear. Therefore, the orange gear must rotate in the opposite direction to the middle blue gear.

Therefore, the correct answer is (B) Counterclockwise.
\end{tcolorbox}

\paragraph{Analysis.}
As shown in Figure~\ref{fig:my_label}, this example requires the direction of motion to be propagated through two successive gear contacts. The OPD baseline correctly recognizes the green arrow and infers that the rightmost gear rotates counterclockwise. It also correctly determines that the middle gear must rotate clockwise. However, it fails to apply the direction reversal at the second contact and consequently predicts that the orange gear also rotates clockwise. The error therefore occurs during relational composition rather than initial visual perception.

S-OPD explicitly follows the two visible gear contacts. Starting from the counterclockwise rotation of the rightmost gear, it reverses the direction once for the middle gear and a second time for the orange gear, yielding the correct answer, counterclockwise. This case illustrates that \textbf{effective visual grounding must be maintained across multiple reasoning steps}: correctly reading the arrow alone is insufficient unless the model consistently applies the spatial interaction represented by each gear contact.

\subsubsection{MMMU-val: Muscle Tissue Image Identification}
\begin{tcolorbox}[enhanced, breakable, frame hidden,
    borderline west={2.5pt}{0pt}{red!70!black},
    colback=red!3, left=5pt, right=3pt, top=3pt, bottom=3pt,
    before skip=5pt, after skip=5pt]
{\bfseries\small OPD baseline {\normalfont(incorrect, answer is A)}}
\smallskip\footnotesize

The image depicts a series of pink, cylindrical structures with a distinct striped or ribbed pattern. These are characteristic of cardiac muscle.

Therefore, the correct answer is: A. Cardiac muscle
\end{tcolorbox}
\begin{tcolorbox}[enhanced, breakable, frame hidden,
    borderline west={2.5pt}{0pt}{blue!70!black},
    colback=blue!3, left=5pt, right=3pt, top=3pt, bottom=3pt,
    before skip=5pt, after skip=5pt]
{\bfseries\small S-OPD (Ours) {\normalfont(correct, answer is C)}}

\smallskip\footnotesize
The image depicts a series of pink, cylindrical structures with a distinct striped or ridged pattern. These are characteristic of muscle tissue, specifically skeletal muscle fibers.

Cardiac muscle is typically arranged in a more complex, branching pattern. Skeletal muscle is composed of long, cylindrical, striated fibers arranged in parallel bundles. The image shows these fibers in a neat, parallel arrangement.

Therefore, the correct answer is C. Skeletal muscle.
\end{tcolorbox}

\paragraph{Analysis.}
Both models detect the prominent striated appearance of the tissue, but they interpret the arrangement of the fibers differently. The OPD baseline associates striation with cardiac muscle and selects option A, without sufficiently accounting for the long, unbranched, and highly parallel organization visible in the image. Its prediction is therefore based on a relevant but non-discriminative visual attribute.

S-OPD considers both the shared attribute and the features that distinguish the candidate tissue types. In particular, it contrasts the branching organization expected of cardiac muscle with the long, cylindrical, parallel fibers characteristic of skeletal muscle. \textbf{This fine-grained comparison leads to the correct selection of option C}. Unlike the preceding examples, which emphasize quantitative and spatial reasoning, this case demonstrates the use of multiple visual attributes for category discrimination.

\subsection{Token-level Probability Visualization and Analysis}
\label{sec:token_visualization}

To examine how teacher gating modulates visually induced policy differences at the token level, we visualize three representative examples covering mathematical reasoning, abstract visual logic, and scientific diagram understanding. For each example, the student first generates a fixed response $y$ from the original image $I$ and the question $q$. We then score the same response and prefix under the original image $I$ and a patch-masked image $\widetilde{I}$ using both the student and teacher models. This controlled comparison isolates how removing visual evidence changes the probability assigned to each generated token.

\begin{figure}[h]
    \centering
    \begin{minipage}{\linewidth}
        \centering
        \refstepcounter{subfigure}
        \label{fig:vis_token_level_a}
        \includegraphics[width=\linewidth]{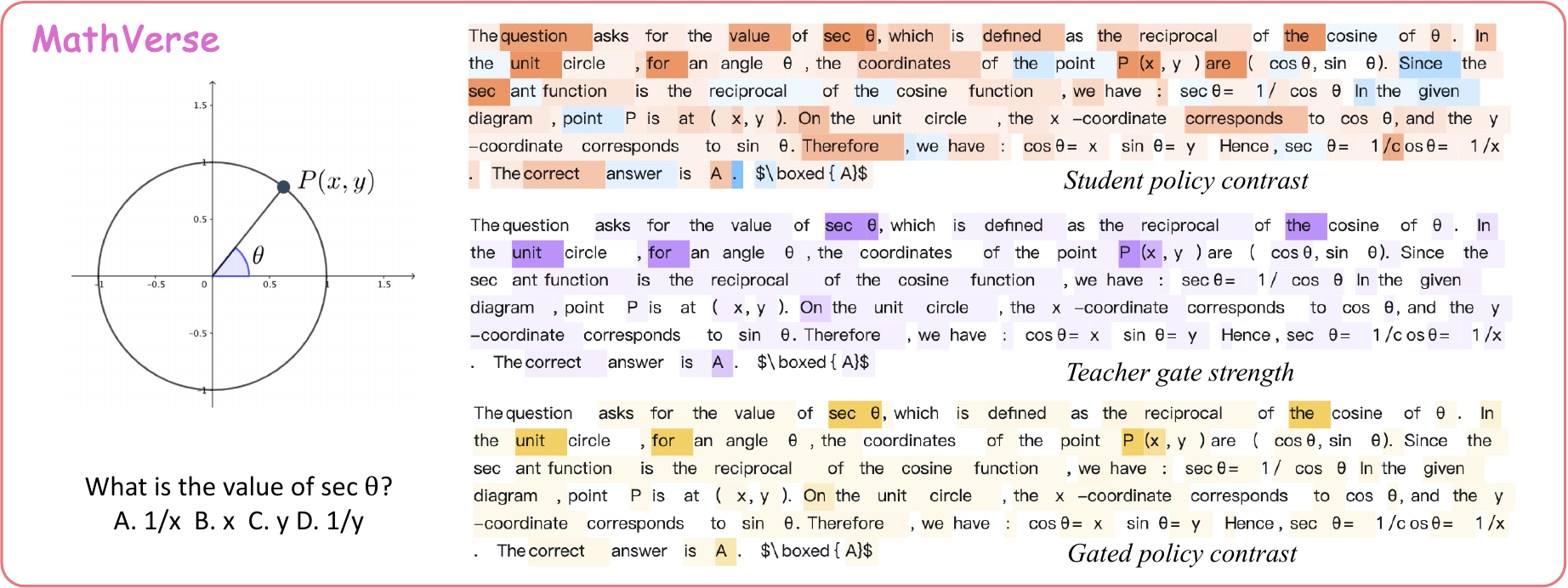}
    \end{minipage}
    \par\vspace{5pt}
    \begin{minipage}{\linewidth}
        \centering
        \refstepcounter{subfigure}
        \label{fig:vis_token_level_b}
        \includegraphics[width=\linewidth]{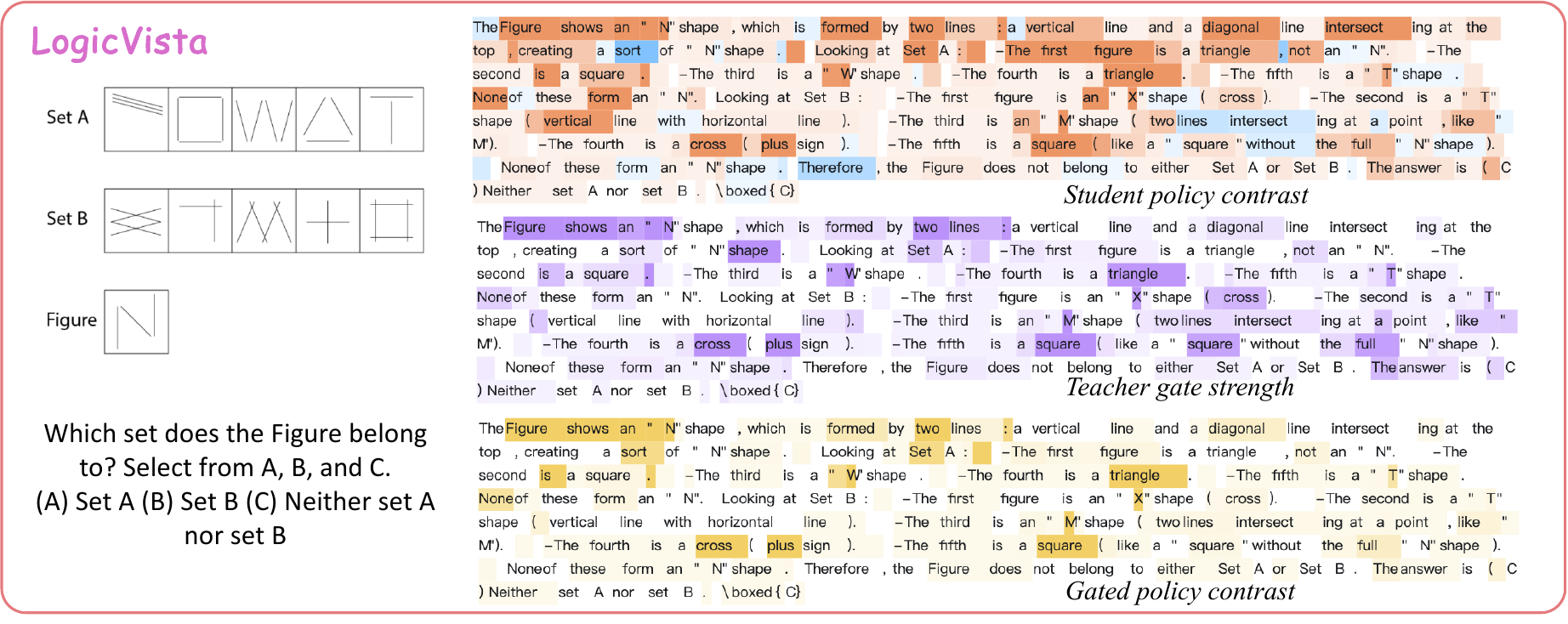}
    \end{minipage}
    \par\vspace{4.5pt}
    \begin{minipage}{\linewidth}
        \centering
        \refstepcounter{subfigure}
        \label{fig:vis_token_level_c}
        \includegraphics[width=\linewidth]{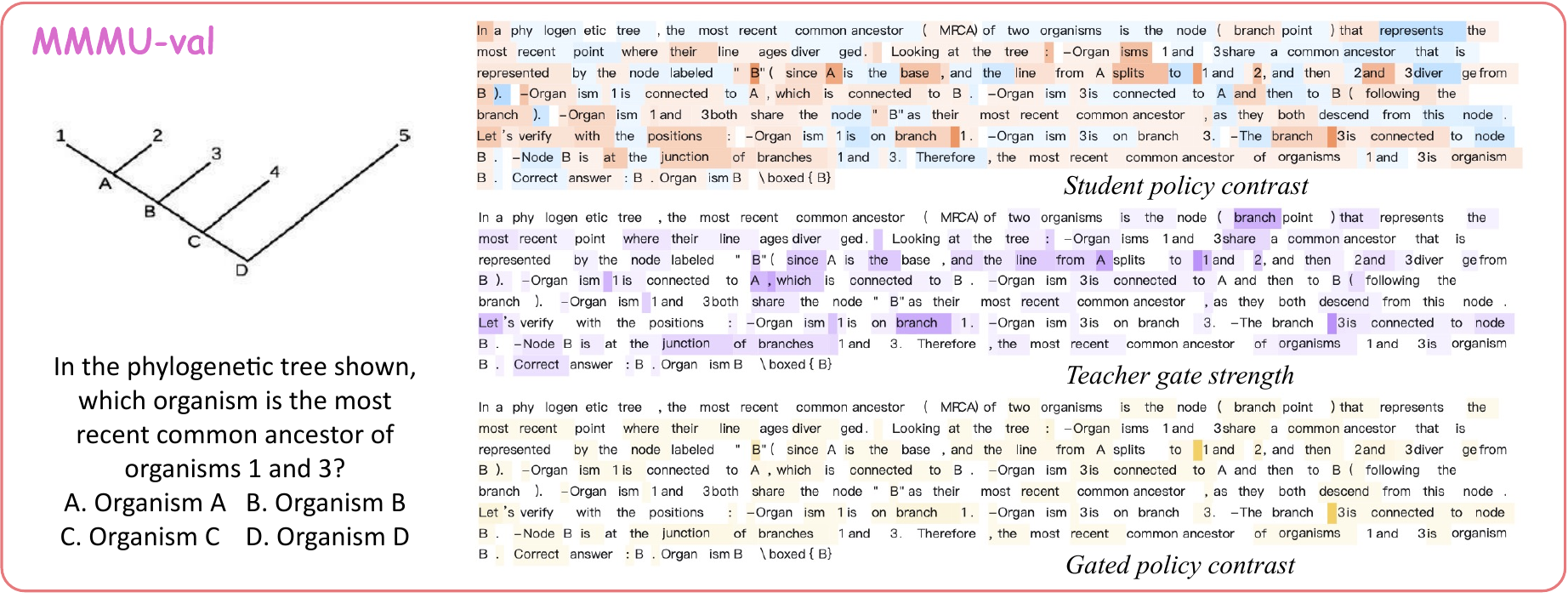}
    \end{minipage}
    \caption{\textbf{Qualitative visualization of token-level policy contrast and teacher gating across three types of visual reasoning}: mathematical geometry (MathVerse), abstract visual logic (LogicVista), and scientific diagram understanding (MMMU-val). In the \textit{student policy contrast}, orange and blue indicate positive and negative changes in token log-probability between the original and masked visual inputs, respectively, while darker colors denote larger absolute changes. For the \textit{teacher gate strength}, darker purple indicates a larger positive teacher log-probability gap between the original and masked images. For the \textit{gated policy contrast}, darker yellow indicates a larger teacher-modulated policy discrepancy. The teacher and student exhibit distinct token-level emphasis across all three tasks, and the gated policy contrast selectively concentrates on key visual elements and decision-relevant reasoning steps, demonstrating the intended effect of our teacher-gating mechanism.
    }
    \label{fig:vis_token_level}
\end{figure}

Figure~\ref{fig:vis_token_level} visualizes the three cases. The signed student policy contrast is defined as
\begin{equation}
    \Delta_t^{S}
    =
    \log p_S(y_t \mid q, I, y_{<t})
    -
    \log p_S(y_t \mid q, \widetilde{I}, y_{<t}),
\end{equation}
where orange indicates that the original image increases the probability of token $y_t$, blue indicates the opposite direction, and color intensity represents the absolute magnitude. The teacher gate strength is
\begin{equation}
    s_t
    =
    \left[
    \log p_T(y_t \mid q, I, y_{<t})
    -
    \log p_T(y_t \mid q, \widetilde{I}, y_{<t})
    \right]_{+},
\end{equation}
with darker purple denoting stronger teacher support from the original visual evidence. Finally, the gated policy contrast combines the teacher gate with the non-negative student contrast:
\begin{equation}
    C_t^{\mathrm{gated}}
    =
    s_t
    \left(
    \exp(d_t)-d_t-1
    \right),
    \qquad
    d_t
    =
    \log p_S(y_t \mid q, \widetilde{I}, y_{<t})
    -
    \log p_S(y_t \mid q, I, y_{<t}).
\end{equation}
Darker yellow therefore marks tokens for which the student policy changes substantially and the teacher simultaneously provides positive visual support.

The three cases reveal consistent but task-dependent behavior. In the MathVerse example, the student contrast responds broadly to visually grounded terms such as ``unit,'' ``sec,'' and the coordinates of $P(x,y)$, whereas the teacher gate places stronger emphasis on mathematical entities such as $\theta$, $\sec\theta$, $P$, and the predicted answer. Their combination concentrates the gated contrast on the quantities required to infer $\sec\theta$ from the unit-circle diagram. In LogicVista, the student is sensitive to several local shape descriptions, while the teacher more selectively emphasizes tokens describing the structural pattern, including intersections and shape categories such as ``cross'',  ``triangle'',  and ``square''. The resulting gated contrast highlights the tokens used to distinguish the two sets. In the MMMU-val example, the contrast is concentrated around branch identities, node relations, and the answer token corresponding to the most recent common ancestor, reflecting the relational structure of the phylogenetic tree.

Across all three task types, the teacher gate and student policy contrast exhibit clearly different token-level emphasis. The teacher signal is therefore not a simple copy or uniform rescaling of the student's visual sensitivity. Instead, the gated policy contrast preserves student discrepancies only when they are supported by the teacher's response to the visual evidence, concentrating the learning signal on visually grounded and decision-relevant reasoning steps. The consistent behavior across distinct visual-reasoning domains supports the intended role of our teacher-gating mechanism as a selective token-level routing strategy.

\end{document}